\documentclass{article}
\usepackage{iclr2027_conference,times}

\usepackage{amsmath,amsfonts,bm}

\def\eqref#1{equation~\ref{#1}}

\def\1{\bm{1}}

\DeclareMathAlphabet{\mathsfit}{\encodingdefault}{\sfdefault}{m}{sl}
\SetMathAlphabet{\mathsfit}{bold}{\encodingdefault}{\sfdefault}{bx}{n}

\usepackage{hyperref}
\usepackage{url}
\usepackage{graphicx}
\usepackage{booktabs}
\usepackage{xcolor}
\usepackage{colortbl} 
\usepackage{multirow}

\iclrfinalcopy
\hypersetup{
  hidelinks,
  pdftitle={Search-E1: Self-Distillation Drives Self-Evolution in Search-Augmented Reasoning},
  pdfauthor={Zihan Liang, Yufei Ma, Ben Chen}
}

\title{Search-E1: Self-Distillation Drives Self-Evolution in Search-Augmented Reasoning}

\author{
\textbf{Zihan Liang$^{*}$ \quad Yufei Ma$^{*}$ \quad Ben Chen$^{*,\dagger}$} \\[3pt]
{\normalfont\small Kuaishou Technology} \\
{\normalfont\footnotesize $^{*}$Equal contribution. \quad $^{\dagger}$Corresponding author.}
}

\begin{document}

\maketitle

\begin{abstract}
Outcome based reinforcement learning provides a simple and effective recipe for training search augmented reasoning agents.
Recent work seeks further improvements through stronger external teachers, auxiliary process supervision, specialized reward shaping, and increasingly elaborate training pipelines.
This raises a natural question: \emph{can agents continue improving by learning from their own successful experience?}
We present \textbf{Search-E1}, a simple self evolution framework that alternates vanilla GRPO with on policy self distillation (OPSD).
After each GRPO round, Search-E1 mines sibling trajectories generated by the policy itself and exposes the search skeleton of an efficient successful trajectory, that is, the sequence of queries it issued, as privileged context.
A token level forward KL objective then aligns the policy under its standard inference context with its own distribution under this privileged context.
In this way, the policy converts contrasts among its own rollouts into dense token level supervision without an external teacher, auxiliary reward model, or annotation beyond reference answers.
On seven single hop and multi hop QA benchmarks, Search-E1 reaches an average Exact Match of $0.440$ with Qwen2.5-3B and $0.482$ with Qwen2.5-7B, the highest seven-benchmark averages in our comparison.
These results show that alternating outcome based exploration with self distillation provides a simple and effective path to iterative improvement in search augmented reasoning.

\end{abstract}

\section{Introduction}
\label{intro}

\par
Large language models have demonstrated strong reasoning capabilities across a wide range of tasks~\citep{deepseek2025r1,openai2024openaio1card,yang2025qwen3}, but their parametric knowledge can be incomplete or outdated.
Search augmented reasoning addresses this limitation by allowing a model to interleave reasoning with calls to an external retriever, deciding when to search and what to query for~\citep{jin2025searchr1,chen2025research,shi2025autorefine,yao2022react}.
Recent work has shown that such agents can be trained effectively with outcome based reinforcement learning such as GRPO~\citep{shao2024deepseekmath}, using final answer correctness as the primary supervision signal.
The resulting recipe is simple: sample trajectories, evaluate their outcomes, and optimize the policy with GRPO.

\par
Recent methods seek further improvements by adding increasingly specialized components, including stronger external supervision, process level reward signals, structured rollout strategies, and manually designed reward shaping~\citep{xu2025thinker,wang2025stepsearch,zhang2025criticsearch,feng2025gigpo,ding2025treegrpo,searchp1}.
These approaches improve search agents from different perspectives, but also introduce additional models, annotations, objectives, or assumptions into the training pipeline.
This raises a natural question: \emph{are these additional components necessary for continued improvement in search augmented reasoning?}

\par
Our starting point is a simple observation.
For the same training question, standard GRPO naturally produces sibling trajectories with substantially different behaviors.
One trajectory may reach the correct answer through a concise and effective search process, while another may issue redundant queries, follow an unproductive reasoning path, or fail entirely.
These contrasts already provide useful supervision for distinguishing more effective behaviors from less effective ones.
Rather than introducing an external supervisor to extract this information, we ask whether the policy can directly learn from the successful experience contained in its own rollouts.

\par
We therefore propose \textbf{Search-E1}, a self evolution framework that alternates GRPO with on policy self distillation (OPSD).
The two stages serve complementary roles.
GRPO explores at the trajectory level and discovers successful behaviors, while OPSD consolidates these discoveries into the policy.
After each GRPO round, Search-E1 mines an efficient successful sibling trajectory and exposes only the queries it issued as privileged context, withholding the retrieved evidence and the final answer.
The same policy then serves as both student and teacher under different conditioning contexts, and a token level forward KL objective transfers the behavior induced by the privileged context back to the standard inference setting.
The resulting training signal is derived from the policy's own rollouts selected using reference answers, without an external teacher, auxiliary reward model, or intermediate annotation.

\par
The exploration and consolidation stages can be iterated for multiple rounds.
Each GRPO round refreshes the on policy trajectory pool, from which successful trajectories are mined to provide privileged supervision for the subsequent OPSD round.
This forms a simple self-evolution loop in which the policy repeatedly learns from experience generated by its own updated behavior.
We further study how the effectiveness of this loop changes across rounds in Section~\ref{sec:self_evolution_exp}.

\par
On seven single hop and multi hop QA benchmarks, Search-E1 reaches an average Exact Match of $0.440$ with Qwen2.5-3B and $0.482$ with Qwen2.5-7B, the highest seven-benchmark averages in our comparison.
Search-E1 improves both single hop and multi hop QA, although it does not lead every task group at both scales.
These results suggest that search agents can be improved iteratively through a simple loop of outcome based reinforcement learning and self distillation, without relying on increasingly elaborate external supervisory machinery.
Our contributions are summarized as follows:

\begin{itemize}
    \item We propose \textbf{Search-E1}, a self evolution framework that alternates vanilla GRPO with on policy self distillation, requiring no external teacher, auxiliary reward model, or intermediate annotation.

    \item We show how differences among sibling trajectories generated by the policy itself can be converted into dense token level supervision, allowing successful behaviors discovered by reinforcement learning to be consolidated and reused in subsequent rounds.

    \item Experiments on seven QA benchmarks at the 3B and 7B scales show consistent improvements over strong outcome reward and process supervision baselines. We further analyze the training dynamics that affect the effectiveness of repeated self evolution across rounds.
\end{itemize}

\section{Method}
\label{method}

\par
\begin{figure}[t]
    \centering
    \includegraphics[width=0.85\linewidth]{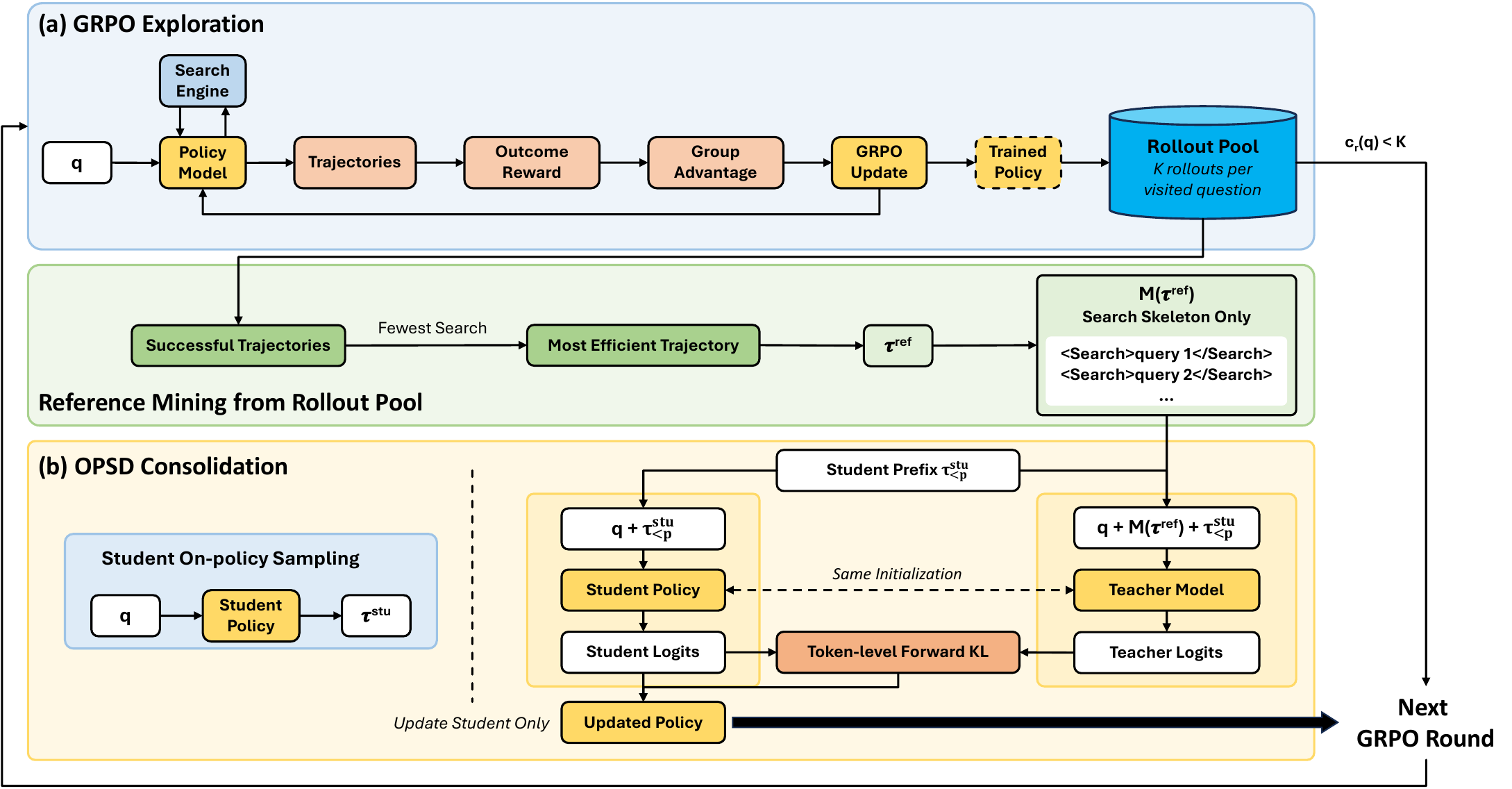}
    \caption{
    Overview of Search-E1.
    GRPO explores with outcome rewards and produces an on-policy rollout pool.
    Successful trajectories provide privileged supervision for OPSD, which consolidates discovered behaviors through token-level self-distillation before the updated policy enters the next GRPO round.
    }
    \label{fig:method}
\end{figure}

Search-E1 alternates two standard learning mechanisms with complementary roles.
GRPO explores new behaviors using outcome rewards, while on policy self distillation consolidates successful behaviors discovered by the policy itself.
We first introduce the search augmented trajectory format and GRPO objective in Section~\ref{sec:prelim}.
We then describe how self generated successful trajectories provide privileged supervision for OPSD in Section~\ref{sec:opsd}.
Finally, Section~\ref{sec:selfevolution} presents the iterative self evolution procedure.
An overview of the complete framework is shown in Figure~\ref{fig:method}.

\subsection{Preliminaries}
\label{sec:prelim}

\paragraph{Search augmented trajectories.}
Following prior work on search augmented reasoning~\citep{jin2025searchr1,shi2025autorefine}, a policy $\pi_\theta$ generates a trajectory $\tau$ by interleaving reasoning with calls to a fixed retriever $\mathcal{E}$.
A trajectory contains \texttt{<think>} spans for internal reasoning, \texttt{<search>} spans for retrieval queries, \texttt{<information>} spans returned by the retriever, and a final \texttt{<answer>} span.
We denote by $\mathcal{A}_\tau$ the positions generated by the policy itself.
Tokens inside \texttt{<information>} spans are excluded from $\mathcal{A}_\tau$ because they are produced by the retriever.

\paragraph{Outcome based GRPO.}
For each question $q$, GRPO samples a group of $G$ trajectories
$\{\tau_1,\ldots,\tau_G\}$ from the rollout policy $\pi_{\theta_{\mathrm{old}}}$.
Each trajectory receives an outcome reward $R_i$ according to normalized EM against the reference answer.
The group normalized advantage is
\begin{equation}
\hat{A}_i =
\frac{
R_i - \operatorname{mean}(\{R_j\}_{j=1}^{G})
}{
\operatorname{std}(\{R_j\}_{j=1}^{G})
}.
\label{eq:grpo_adv}
\end{equation}
The policy is optimized using the standard GRPO objective
\begin{equation}
\mathcal{L}_{\mathrm{GRPO}}(\theta)
=
-\mathbb{E}
\left[
\min
\left(
r_p(\theta)\hat{A}_i,
\operatorname{clip}
\left(
r_p(\theta),1-\epsilon,1+\epsilon
\right)\hat{A}_i
\right)
\right],
\label{eq:grpo}
\end{equation}
where
\begin{equation}
r_p(\theta)
=
\pi_\theta(a_p\mid\tau_{<p})
/
\pi_{\theta_{\mathrm{old}}}(a_p\mid\tau_{<p})
.
\end{equation}

Search-E1 leaves this stage unchanged.
In particular, we introduce no process reward, auxiliary critic, or modification to the GRPO advantage.

\subsection{Self Distillation from Self Generated Privileged Trajectories}
\label{sec:opsd}

\par
The key observation behind Search-E1 is that GRPO already generates the supervision required for further improvement.
For the same question, some trajectories solve the problem through an effective search process while others fail or take less efficient paths.
We use the successful trajectories generated by the policy itself as privileged information for on policy self distillation.

\paragraph{Mining privileged trajectories.}
After GRPO round $r$, we sample $K$ trajectories
$\{\tau_q^{(1)},\ldots,\tau_q^{(K)}\}$
for each question visited during that round using the resulting policy $\pi_r^{\mathrm{RL}}$.
Each trajectory is labeled by its final answer correctness against the reference answer.
Among the successful trajectories, we select an efficient trajectory as the privileged reference $\tau_q^{\mathrm{ref}}$.
By default, we prefer the trajectory with the fewest search calls and break ties by the total length of its search skeleton.
Questions for which no successful trajectory is observed do not provide a privileged reference for the current OPSD round.
Importantly, $\tau_q^{\mathrm{ref}}$ is generated entirely by the current policy.
It is neither a human annotated reasoning trace nor a trajectory synthesized by a stronger model.

\paragraph{Search skeleton as privileged context.}
The privileged reference enters the teacher context only through its search skeleton.
We define a masking operator $\mathcal{M}$ that retains the \texttt{<search>} spans of a trajectory in their original order and discards all remaining content, including \texttt{<think>} spans, retrieved \texttt{<information>} passages, and the final \texttt{<answer>} span.
The teacher therefore observes which queries were issued and in what order, but neither the evidence they returned nor the answer they supported.
This restriction is deliberate.
Retaining the retrieved passages or the final answer would make the answer directly predictable from the teacher context, so the teacher would place mass on continuations that skip retrieval and would push the student toward answering without evidence.
The skeleton also keeps the privileged context short, since retrieved passages dominate the length of a search augmented trajectory.
We isolate the effect of this choice in Section~\ref{sec:ablation} and analyze it further in Appendix~\ref{app:privileged_context}.

\paragraph{On policy student sampling.}
Given a question $q$ with privileged reference $\tau_q^{\mathrm{ref}}$, the student samples its own trajectory
\begin{equation}
\tau^{\mathrm{stu}}
\sim
\pi_\theta(\cdot\mid P_S(q)),
\label{eq:student_rollout}
\end{equation}
where $P_S(q)$ is the standard inference prompt containing the task instruction and question.
Thus, the distillation objective is evaluated on trajectories generated by the policy being optimized, rather than on the privileged reference itself.

\paragraph{Privileged teacher conditioning.}
The student and teacher originate from the same policy but receive different information.
The student observes only the standard inference context,
\begin{equation}
P_S(q)
=
[\text{instruction}] \,\|\, q,
\end{equation}
while the teacher additionally observes the search skeleton of the self generated privileged reference,
\begin{equation}
P_T(q,\tau_q^{\mathrm{ref}})
=
[\text{instruction}] \,\|\, q \,\|\, \mathcal{M}(\tau_q^{\mathrm{ref}}).
\end{equation}
Both policies evaluate the same student generated trajectory $\tau^{\mathrm{stu}}$.
For each policy generated position $p\in\mathcal{A}_{\tau^{\mathrm{stu}}}$, we define
\begin{align}
P_p^{\mathrm{stu}}
&=
\pi_{\theta}
\left(
\cdot
\mid
P_S(q),
\tau_{<p}^{\mathrm{stu}}
\right),\\
P_p^{\mathrm{tch}}
&=
\operatorname{sg}
\left[
\pi_r^{\mathrm{RL}}
\left(
\cdot
\mid
P_T(q,\tau_q^{\mathrm{ref}}),
\tau_{<p}^{\mathrm{stu}}
\right)
\right],
\end{align}
where $\operatorname{sg}[\cdot]$ denotes stop gradient.
At the beginning of each OPSD round, both teacher and student are initialized from $\pi_r^{\mathrm{RL}}$.
The teacher is held fixed during distillation, while the student is updated.
The distinction between the two therefore arises from privileged conditioning rather than from a stronger external model.

\paragraph{Token level distillation.}
Following OPSD~\citep{zhao2026opsd}, we minimize a forward KL divergence between teacher and student along the student's own rollout.
We use pointwise clipping to prevent a small number of large vocabulary level divergences from dominating the update:
\begin{equation}
\mathcal{L}_{\mathrm{OPSD}}
=
\frac{1}{
|\mathcal{A}_{\tau^{\mathrm{stu}}}|
}
\sum_{p\in\mathcal{A}_{\tau^{\mathrm{stu}}}}
\sum_{v\in\mathcal{V}}
\min
\left(
P_p^{\mathrm{tch}}(v)
\log
\frac{
P_p^{\mathrm{tch}}(v)
}{
P_p^{\mathrm{stu}}(v)
},
\tau_{\mathrm{clip}}
\right).
\label{eq:opsd}
\end{equation}
The teacher never generates the student response.
It only provides a privileged next token distribution over the student's on policy trajectory.
Consequently, the student receives dense token level supervision while remaining on its own generation distribution.
At inference time, the privileged context and the teacher evaluation are removed.

\subsection{Iterative Self Evolution}
\label{sec:selfevolution}

Search-E1 alternates GRPO exploration and OPSD consolidation over multiple rounds.
Suppose round $r$ starts from policy $\pi_r$.
GRPO first optimizes $\pi_r$ using only outcome rewards, producing $\pi_r^{\mathrm{RL}}$.
We then sample a rollout pool from $\pi_r^{\mathrm{RL}}$, mine successful privileged trajectories, and apply OPSD to obtain the policy $\pi_{r+1}$.
The resulting policy enters the next GRPO round.
The rollout pool also determines which questions remain useful for subsequent exploration.
Let $\mathcal{V}_r \subseteq \mathcal{D}_r$ denote the questions visited during round $r$, and let $c_r(q)$ denote the number of correct trajectories among the $K$ rollouts sampled from $\pi_r^{\mathrm{RL}}$ for a question $q \in \mathcal{V}_r$.

For OPSD, we use questions for which at least one successful trajectory is available:
\begin{equation}
\mathcal{D}^{\mathrm{OPSD}}_r
=
\left\{
q \in \mathcal{V}_r : c_r(q) > 0
\right\}.
\label{eq:opsd_data}
\end{equation}
For the next GRPO round, we retain questions that are not uniformly solved:
\begin{equation}
\mathcal{D}_{r+1}
=
\mathcal{D}_r \setminus
\left\{
q \in \mathcal{V}_r : c_r(q) = K
\right\}.
\label{eq:grpo_recycling}
\end{equation}
All failure questions remain for the next GRPO round but provide no OPSD reference; mixed outcome questions participate in both stages; uniformly solved questions may provide OPSD supervision in the current round but are removed from the next GRPO round.

The resulting procedure forms a closed self evolution loop with fixed reference answers and a fixed verifier.
GRPO discovers successful behaviors through exploration, the policy's own rollout pool supplies privileged supervision, and OPSD consolidates these behaviors before the next round begins.
No external teacher, process reward model, or new annotation is introduced between rounds.

\section{Experiments}
\label{experiments}

\par
We evaluate Search-E1 with four questions.
First, does the simple alternation of GRPO and OPSD improve over existing search augmented reasoning methods?
Second, do the gains accumulate across multiple self evolution rounds?
Third, what determines whether an OPSD updated policy remains suitable for further GRPO exploration?
Finally, which components of trajectory mining and data recycling are necessary for the observed improvements?

\subsection{Experimental Setup}
\label{sec:exp_setup}

\paragraph{Datasets and evaluation.}
Following the evaluation protocol of prior search augmented reasoning work~\citep{jin2025searchr1,shi2025autorefine}, we train on the combined training sets of Natural Questions (NQ)~\citep{kwiatkowski2019naturalquestions} and HotpotQA~\citep{yang2018hotpotqa}, containing approximately 170K question answer pairs.
We evaluate on seven QA benchmarks, including three single hop datasets, NQ, TriviaQA~\citep{joshi2017triviaqa}, and PopQA~\citep{mallen2023popqa}, and four multi hop datasets, HotpotQA, 2WikiMultihopQA (2Wiki)~\citep{ho20202wiki}, MuSiQue~\citep{trivedi2022musique}, and Bamboogle~\citep{press2023bamboogle}.
Performance is measured by Exact Match (EM) after standard answer normalization, including lowercasing, punctuation removal, and article removal.
Detailed dataset statistics are provided in Appendix~\ref{app:datasets}.

\paragraph{Baselines.}
We compare Search-E1 with retrieval free baselines, single retrieval methods, and multi round search augmented reasoning methods.
Retrieval free baselines include Direct Generation, supervised fine tuning, and R1 style reinforcement learning~\citep{deepseek2025r1}.
We include Naive RAG~\citep{lewis2020naiverag} as a single retrieval baseline and Search-o1~\citep{li2025searcho1} and IRCoT~\citep{trivedi2023ircot} as inference time search methods.
For reinforcement learning based search agents, we compare against Search-R1~\citep{jin2025searchr1}, ReSearch~\citep{chen2025research}, AutoRefine~\citep{shi2025autorefine}, and MR-Search~\citep{xiao2025mrsearch}.
We further include methods that introduce denser supervision or additional training structure, including StepSearch~\citep{wang2025stepsearch}, GiGPO~\citep{feng2025gigpo}, Thinker~\citep{xu2025thinker}, SD-Search~\citep{ma2026sdsearch}, and IG-Search~\citep{liang2026igsearch}.
Baseline results follow published reports or our AutoRefine-Instruct (7B) reproduction; Appendix~\ref{app:baseline_sources} details sources and configurations.

\paragraph{Models and retrieval.}
We evaluate both Base and Instruct variants of Qwen2.5 at the 3B and 7B scales~\citep{qwen2025qwen25technicalreport}.
Our Search-E1 runs and internal controls use the same fixed retrieval environment.
The retriever is E5-base-v2~\citep{wang2024e5} over a December 2018 Wikipedia corpus and returns the top three passages for each query.
These runs allow at most four search calls; external baselines retain their reported limits.
We use a rollout temperature of 1.0 throughout training and evaluation.

\paragraph{Training configuration.}
Search-E1 is implemented with veRL~\citep{Sheng2025hybridflow}.
GRPO uses an Exact Match outcome reward with group size $G=5$, learning rate $1\times10^{-6}$, KL coefficient $\beta=0.001$, and clip ratio $\epsilon=0.2$.
After each GRPO round, we sample $K=5$ trajectories for each question visited during that round to construct the rollout pool used for privileged trajectory mining and data recycling, so mining and the recycling rules of Equations~\ref{eq:opsd_data} and~\ref{eq:grpo_recycling} are applied to the questions actually visited in round $r$.
OPSD uses forward KL with pointwise clipping at $\tau_{\mathrm{clip}}=10$.
The privileged context contains only the search skeleton of the reference trajectory and is truncated to 512 tokens.
The distillation loss is applied at all policy generated positions of the student rollout.
For the transition to the next GRPO round, we select the OPSD checkpoint with the highest validation EM among checkpoints whose mean token entropy retains at least $85\%$ of its value at the beginning of the OPSD round.
We run three GRPO and OPSD cycles for the 3B models and two cycles for the 7B models.
Unless otherwise specified, ablations are conducted with Qwen2.5-3B-Instruct.
Complete hyperparameters are reported in Appendix~\ref{app:hyperparams}.
Appendix~\ref{app:additional_experiments} reports additional controls, search budgets, retrieval backends, and backbones.

\subsection{Main Results}
\label{sec:main}

\begin{table}[!htbp]
\centering
\caption{
Exact Match on seven QA benchmarks with Qwen2.5-3B.
\textbf{Bold} denotes the best result and \underline{underline} the second best in each column.
``--'' denotes results not reported by the original paper.
}
\label{tab:main_3b}
\setlength{\tabcolsep}{2.5pt}
\footnotesize
\begin{tabular}{l ccc cccc c}
\toprule
& \multicolumn{3}{c}{\textit{Single-Hop}} &
\multicolumn{4}{c}{\textit{Multi-Hop}} & \\
\cmidrule(lr){2-4} \cmidrule(lr){5-8}
\textbf{Method}
& NQ & TriviaQA & PopQA
& HotpotQA & 2Wiki & MuSiQue & Bamboogle
& \textbf{Avg.} \\
\midrule

\multicolumn{9}{c}{\textit{Outcome Reward and Standard Search Training}} \\
\midrule
Search-R1-Base
& 0.421 & 0.583 & 0.413 & 0.297 & 0.274 & 0.066 & 0.128 & 0.312 \\
Search-R1-Instruct
& 0.397 & 0.565 & 0.391 & 0.331 & 0.310 & 0.124 & 0.232 & 0.336 \\
ReSearch-Base
& 0.427 & 0.597 & 0.430 & 0.305 & 0.272 & 0.074 & 0.128 & 0.319 \\
ReSearch-Instruct
& 0.365 & 0.571 & 0.395 & 0.351 & 0.272 & 0.095 & 0.266 & 0.331 \\
AutoRefine-Base
& 0.467 & 0.620 & 0.450 & 0.405 & 0.393 & 0.157 & 0.344 & 0.405 \\
AutoRefine-Instruct
& 0.436 & 0.597 & 0.447 & 0.404 & 0.380 & 0.169 & 0.336 & 0.396 \\
MR-Search-Base
& \underline{0.477} & 0.635 & 0.460 & 0.419 & 0.401 & 0.165 & 0.344 & 0.414 \\
\midrule

\multicolumn{9}{c}{\textit{Additional Process or Step Level Supervision}} \\
\midrule
StepSearch-Base
& -- & -- & -- & 0.329 & 0.339 & 0.181 & 0.328 & -- \\
StepSearch-Instruct
& -- & -- & -- & 0.345 & 0.320 & 0.174 & 0.344 & -- \\
GiGPO-Instruct
& 0.420 & 0.595 & 0.424 & 0.369 & 0.370 & 0.126 & \textbf{0.641} & 0.421 \\
Thinker-Instruct
& 0.439 & 0.598 & 0.469 & 0.400 & \textbf{0.469} & \textbf{0.214} & 0.424 & 0.430 \\
SD-Search-Base
& 0.470 & 0.624 & 0.467 & 0.425 & 0.420 & 0.188 & 0.402 & 0.428 \\
SD-Search-Instruct
& 0.469 & 0.624 & 0.465 & 0.424 & 0.419 & 0.184 & 0.404 & 0.427 \\
IG-Search-Base
& 0.476 & \underline{0.637} & \underline{0.476} & \textbf{0.436} & 0.415 & 0.179 & 0.390 & 0.430 \\
IG-Search-Instruct
& 0.450 & 0.614 & 0.460 & 0.431 & \underline{0.440} & 0.191 & 0.424 & 0.430 \\
\midrule

\textbf{Search-E1-Base (Ours)}
& \textbf{0.479} & \textbf{0.641} & \textbf{0.478}
& \underline{0.432} & 0.429 & 0.190 & 0.416
& \underline{0.438} \\

\textbf{Search-E1-Instruct (Ours)}
& 0.474 & 0.626 & 0.461
& 0.427 & 0.436 & \underline{0.193} & \underline{0.464}
& \textbf{0.440} \\

\bottomrule
\end{tabular}
\end{table}

\par
Table~\ref{tab:main_3b} reports results at the 3B scale.
At the 3B scale, Search-E1-Instruct achieves the highest average EM of $0.440$, while Search-E1-Base reaches $0.438$.
Both variants outperform the strongest outcome reward baseline, MR-Search-Base ($0.414$), by more than two points on average.
They also improve over the strongest variants of SD-Search and IG-Search, which reach average EM scores of $0.428$ and $0.430$, respectively.
These results indicate that repeated alternation between standard outcome reward RL and self distillation can outperform methods that introduce additional step level objectives, despite leaving the underlying GRPO objective unchanged.

\par
The two Search-E1 variants show complementary behavior.
The Base model performs better on the single hop benchmarks, whereas the Instruct model is stronger on several multi hop datasets, including 2Wiki, MuSiQue, and Bamboogle.
Among the Instruct variants, Search-E1 has the highest single hop average at both scales and the highest multi hop average at 3B. At 7B, its multi hop average is below GiGPO and slightly below IG-Search.

\begin{table}[!htbp]
\centering
\caption{
Exact Match on seven QA benchmarks with Qwen2.5-7B.
\textbf{Bold} denotes the best result and \underline{underline} the second best in each column.
``--'' denotes results not reported by the original paper.
}
\label{tab:main_7b}
\setlength{\tabcolsep}{2.5pt}
\footnotesize
\begin{tabular}{l ccc cccc c}
\toprule
& \multicolumn{3}{c}{\textit{Single-Hop}} &
\multicolumn{4}{c}{\textit{Multi-Hop}} & \\
\cmidrule(lr){2-4} \cmidrule(lr){5-8}
\textbf{Method}
& NQ & TriviaQA & PopQA
& HotpotQA & 2Wiki & MuSiQue & Bamboogle
& \textbf{Avg.} \\
\midrule

\multicolumn{9}{c}{\textit{Outcome Reward and Standard Search Training}} \\
\midrule
Search-R1-Base
& 0.469 & 0.627 & 0.449 & 0.410 & 0.272 & 0.173 & 0.456 & 0.408 \\
Search-R1-Instruct
& 0.429 & 0.623 & 0.427 & 0.386 & 0.346 & 0.162 & 0.400 & 0.396 \\
AutoRefine-Base
& 0.484 & 0.659 & 0.487 & 0.451 & 0.405 & 0.187 & 0.512 & 0.455 \\
AutoRefine-Instruct
& 0.473 & 0.652 & 0.475 & 0.443 & 0.398 & 0.181 & 0.496 & 0.445 \\
MR-Search-Base
& \underline{0.502} & 0.666 & 0.472 & 0.468 & 0.436 & 0.221 & 0.452 & 0.460 \\
\midrule

\multicolumn{9}{c}{\textit{Additional Process or Step Level Supervision}} \\
\midrule
StepSearch-Base
& -- & -- & -- & 0.380 & 0.385 & 0.216 & 0.467 & -- \\
StepSearch-Instruct
& -- & -- & -- & 0.386 & 0.366 & \underline{0.226} & 0.400 & -- \\
GiGPO-Instruct
& 0.464 & 0.647 & 0.461 & 0.416 & 0.436 & 0.189 & \textbf{0.689} & 0.472 \\
Thinker-Instruct
& 0.450 & 0.642 & 0.484 & 0.421 & \textbf{0.469} & 0.221 & 0.480 & 0.452 \\
SD-Search-Base
& 0.500 & 0.665 & 0.489 & 0.468 & 0.438 & 0.204 & 0.532 & 0.471 \\
SD-Search-Instruct
& 0.495 & 0.668 & 0.494 & 0.471 & 0.441 & 0.222 & 0.544 & 0.476 \\
IG-Search-Base
& 0.501 & 0.675 & \underline{0.504} & \underline{0.473} & 0.430 & 0.205 & 0.544 & 0.476 \\
IG-Search-Instruct
& 0.490 & 0.665 & 0.495 & \textbf{0.480} & 0.445 & 0.215 & \underline{0.560} & 0.479 \\
\midrule

\textbf{Search-E1-Base (Ours)}
& \textbf{0.503} & \underline{0.683} & \textbf{0.505}
& 0.472 & 0.447 & 0.225 & 0.525
& \underline{0.480} \\

\textbf{Search-E1-Instruct (Ours)}
& 0.487 & \textbf{0.694} & 0.497
& 0.469 & \underline{0.455} & \textbf{0.236} & 0.534
& \textbf{0.482} \\

\bottomrule
\end{tabular}
\end{table}

\par
Table~\ref{tab:main_7b} reports the corresponding results at the 7B scale.
At the 7B scale, Search-E1 shows the same overall pattern.
Search-E1-Instruct reaches an average EM of $0.482$, while the Base variant reaches $0.480$.
Both outperform the strongest standard outcome reward baseline, MR-Search-Base ($0.460$), as well as the strongest compared method with additional step level supervision, IG-Search-Instruct ($0.479$).
Both Search-E1 variants improve over their corresponding SD-Search and IG-Search counterparts.
Search-E1-Base reaches $0.480$, compared with $0.471$ for SD-Search-Base and $0.476$ for IG-Search-Base, while Search-E1-Instruct reaches $0.482$, compared with $0.476$ and $0.479$, respectively.
The relative strengths of the Base and Instruct variants remain consistent with the 3B results.
The Base model performs particularly well on the single hop benchmarks, whereas the Instruct model attains higher performance on 2Wiki and MuSiQue.
Importantly, the improvement of Search-E1 persists as the backbone grows from 3B to 7B, suggesting that the benefit of self generated privileged supervision is not restricted to smaller policies.

\subsection{Self Evolution Across Rounds}
\label{sec:self_evolution_exp}

\par
\begin{figure}[!htbp]
\centering

\includegraphics[width=0.55\linewidth]{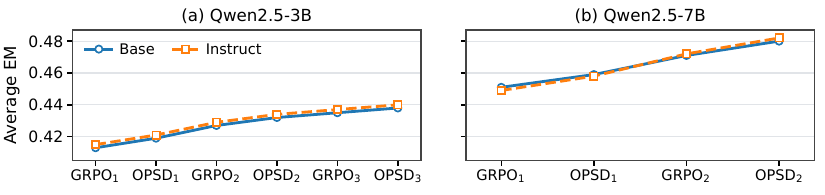}

\caption{
Stage wise average Exact Match during Search-E1 training.
Each cycle alternates GRPO exploration with OPSD consolidation.
Performance improves after both stages across successive cycles, while the magnitude of improvement gradually decreases in later rounds.
}
\label{fig:self_evolution}
\end{figure}

The final results establish the overall effectiveness of Search-E1, but do not reveal how the improvement accumulates across successive rounds.
We therefore evaluate the policy after every GRPO and OPSD stage.
Figure~\ref{fig:self_evolution} reports the average EM throughout the complete self evolution process.

\par
At the 3B scale, both Base and Instruct models improve consistently over three cycles.
Search-E1-Base increases from $0.413$ after the first GRPO round to $0.438$ after the third OPSD round, while Search-E1-Instruct improves from $0.415$ to $0.440$.
Importantly, the gains are not confined to the distillation stages.
Each OPSD round improves upon the preceding GRPO checkpoint, and the subsequent GRPO round is able to further improve from the distilled policy.
For example, Search-E1-Instruct improves from $0.415$ to $0.421$ after the first OPSD stage and then reaches $0.429$ after the following GRPO round.
This alternating improvement indicates that OPSD not only consolidates successful behaviors discovered by GRPO, but also preserves sufficient room for subsequent outcome based exploration.

\par
The gains gradually diminish in later cycles.
For Search-E1-Instruct, the improvement from the first GRPO checkpoint to the first OPSD checkpoint is $0.006$, whereas the final OPSD stage adds $0.003$.
A similar trend is observed for the Base model.
This suggests that repeated self evolution approaches saturation rather than producing unstable or unbounded improvement.

\par
The same pattern persists at the 7B scale.
Search-E1-Base improves from $0.451$ after the first GRPO round to $0.480$ after two complete cycles, while Search-E1-Instruct improves from $0.449$ to $0.482$.
These results show that alternating outcome based exploration with self distillation provides cumulative gains across rounds at both model scales.
Per benchmark results for every stage of all four backbones are reported in Appendix~\ref{app:full_results}.

\begin{figure*}[!htbp]
\centering

\includegraphics[width=0.55\linewidth]{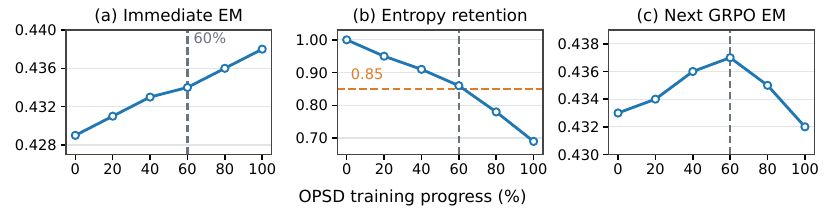}

\caption{
Effect of OPSD checkpoint selection on Qwen2.5-3B-Instruct.
As OPSD progresses, immediate validation EM continues to improve, while policy entropy steadily decreases.
However, initializing the next GRPO round from increasingly late OPSD checkpoints reduces downstream performance.
The vertical dashed line marks the checkpoint selected by our entropy aware handoff rule, while the horizontal dashed line in panel (b) denotes the $0.85$ entropy retention threshold.
}
\label{fig:evolvability}
\end{figure*}

\subsection{Preserving Evolvability Across Rounds}
\label{sec:evolvability}

\par
The previous results show that GRPO and OPSD can produce cumulative improvements across multiple cycles.
A natural question is how to choose the OPSD checkpoint that initializes the next GRPO round.
Selecting the checkpoint with the highest immediate validation accuracy appears reasonable, but this criterion ignores whether the resulting policy remains sufficiently diverse for subsequent exploration.

\par
We study this effect by evaluating multiple checkpoints from the second OPSD round of Qwen2.5-3B-Instruct.
For each checkpoint, we measure its immediate validation EM, its mean token entropy over policy generated positions on a fixed calibration set relative to the beginning of the OPSD round, and the final EM obtained after using it to initialize an otherwise identical GRPO round.
Figure~\ref{fig:evolvability} summarizes the results.

\par
Immediate validation performance improves steadily as OPSD progresses, increasing from $0.429$ before distillation to $0.438$ at the final checkpoint.
This improvement is accompanied by a substantial reduction in policy entropy.
The entropy retention ratio decreases from $1.00$ to $0.69$, indicating that extended distillation progressively sharpens the policy distribution.

\par
This sharpening improves current performance but eventually becomes unfavorable for further reinforcement learning.
The subsequent GRPO round reaches its highest EM of $0.437$ when initialized from an intermediate OPSD checkpoint with an entropy retention of $0.86$.
Continuing OPSD increases immediate EM to $0.438$, but reduces the downstream GRPO result to $0.432$.
Thus, the checkpoint that performs best immediately after distillation is not necessarily the checkpoint from which the policy can improve most in the next round.

\par
This observation motivates the entropy aware handoff rule used in Search-E1.
Among OPSD checkpoints that retain at least $85\%$ of the entropy at the beginning of the round, we select the checkpoint with the highest validation EM.
The rule preserves the performance benefit of self distillation while avoiding excessive distribution sharpening that can limit subsequent exploration.
Additional analysis in Appendix~\ref{app:entropy_diversity} shows that the entropy reduction is accompanied by lower rollout diversity, including fewer distinct trajectories and smaller pairwise trajectory differences.
Appendix~\ref{app:handoff} sweeps the retention threshold and Appendix~\ref{app:behavior} verifies that the gain does not come from a change in retrieval.

\subsection{Ablation Studies}
\label{sec:ablation}

\begin{table}[!htbp]
\centering
\caption{
Ablation study on Qwen2.5-3B-Instruct.
Single and Multi denote the average Exact Match over the three single hop and four multi hop benchmarks, respectively.
}
\label{tab:ablation}
\setlength{\tabcolsep}{5pt}
\footnotesize
\begin{tabular}{lccc}
\toprule
\textbf{Configuration}
& \textbf{Single}
& \textbf{Multi}
& \textbf{Avg.} \\
\midrule

\textbf{Search-E1}
& \textbf{0.520}
& \textbf{0.380}
& \textbf{0.440} \\

\midrule
\multicolumn{4}{l}{\textit{Iterative Training}} \\
\midrule

GRPO only
& 0.507 & 0.365 & 0.426 \\

SFT on mined trajectories
& 0.512 & 0.365 & 0.428 \\

Distillation once after all GRPO rounds
& 0.514 & 0.370 & 0.432 \\

Reuse first round reference pool
& 0.511 & 0.369 & 0.430 \\

w/o entropy aware handoff
& 0.513 & 0.375 & 0.434 \\

\midrule
\multicolumn{4}{l}{\textit{Trajectory and Data Selection}} \\
\midrule

Full trajectory as privileged context
& 0.506 & 0.356 & 0.420 \\

Random successful reference
& 0.516 & 0.374 & 0.435 \\

Longest successful reference
& 0.510 & 0.372 & 0.431 \\

w/o question recycling
& 0.510 & 0.374 & 0.432 \\

\bottomrule
\end{tabular}
\end{table}

\par
Table~\ref{tab:ablation} isolates the main design choices of Search-E1.
Unless otherwise specified, all variants use the same three cycle schedule, the same question pool, and Qwen2.5-3B-Instruct.
Appendix~\ref{app:consolidation} compares further alternatives to the consolidation stage, and Appendix~\ref{app:objective} varies the divergence and the alignment scope.

\paragraph{Is alternating self distillation necessary?}
Removing OPSD and running GRPO alone under the same three round schedule reduces the average EM from $0.440$ to $0.426$.
This shows that the improvement across rounds cannot be explained simply by allocating additional outcome reward optimization.
The effect is particularly pronounced on the multi hop benchmarks, where the average decreases from $0.380$ to $0.365$.
Since Search-E1 executes more optimizer steps in total than this variant, Appendix~\ref{app:budget} further reports a control that runs GRPO alone for the full number of steps executed by Search-E1, which saturates at $0.427$ and then declines.

\par
The form of consolidation also matters.
Replacing OPSD with supervised fine tuning on the same mined trajectories, which is the most direct way of reusing self generated experience, reaches only $0.428$.
Imitating the reference sequence recovers a small part of the benefit, whereas aligning the student to the teacher distribution along its own rollout supplies supervision at every position the student actually visits.
Applying a single distillation stage after three GRPO rounds, rather than alternating, reaches $0.432$.
Consolidation is therefore not simply a post training step.
Its value depends on being interleaved with exploration, so that every round of reinforcement learning starts from a policy that has already absorbed the behaviors discovered in the previous one.

\paragraph{How should self generated experience be used?}
What the teacher is allowed to see matters more than any other choice we examined.
Passing the full reference trajectory, including its retrieved passages and final answer, reduces the average EM to $0.420$, which is below the $0.426$ obtained without any distillation at all.
A teacher that can already read the answer from its own context assigns low probability to issuing further queries, so the student is pushed toward answering on weaker evidence.
Appendix~\ref{app:privileged_context} shows that this variant also reduces the number of search calls per rollout from $2.12$ to $1.38$ and lowers the fraction of retrievals that return the gold answer.
Restricting the privileged context to the search skeleton removes this failure mode.
Appendix~\ref{app:case_study} traces a single question through the schedule to show what this change looks like on one rollout.

\par
Among skeletons, which successful trajectory is selected still matters.
Replacing the efficient successful trajectory with a randomly selected successful trajectory reduces average EM to $0.435$, while selecting the longest successful trajectory further decreases performance to $0.431$.
Successful trajectories therefore do not provide equally useful supervision.
Favoring solutions that reach the correct answer with fewer search calls provides a cleaner privileged context for distillation.

\par
Finally, retaining every visited question for the next exploration round, rather than removing those that all rollouts solve, reduces performance from $0.440$ to $0.432$.
A question that every rollout answers correctly yields a zero normalized advantage, so it consumes rollout budget without contributing gradient, and the share of such questions grows with every round.
Removing them concentrates later exploration on the failure and mixed outcome questions, where search behavior can still change.
Appendix~\ref{app:data_recycling} compares alternative recycling rules and Appendix~\ref{app:reference_selection} compares alternative reference selection rules.

\section{Conclusion}

\par
We present Search-E1, a simple self evolution framework for search augmented reasoning that alternates vanilla GRPO with on policy self distillation.
By using successful trajectories generated by the policy itself as privileged supervision, Search-E1 enables iterative improvement without external teachers, process reward models, or annotations beyond reference answers.
Across seven QA benchmarks, Search-E1 achieves average EM scores of $0.440$ at the 3B scale and $0.482$ at the 7B scale, the highest seven-benchmark averages in our comparison.
Our results suggest that alternating outcome based exploration with self distillation provides a simple and effective path toward self evolving search agents.

\section*{AI use statement}

We used AI writing tools to help polish the language of the manuscript.

\bibliography{iclr2027_conference}
\bibliographystyle{iclr2027_conference}

\clearpage
\appendix

\section{Related Work}
\label{app:related}

\subsection{Reinforcement Learning for Search Augmented Reasoning}

Retrieval augmented generation equips language models with external knowledge beyond their parametric memory~\citep{lewis2020naiverag,borgeaud2022retro,izacard2022atlas}.
Early approaches typically retrieve before generation or interleave retrieval with reasoning through prompting~\citep{yao2022react,trivedi2023ircot,li2025searcho1}.
More recent work trains the model itself to decide when and how to search.
Search-R1~\citep{jin2025searchr1} establishes a simple formulation in which search calls are embedded directly into the reasoning trajectory and the policy is optimized with GRPO using final answer correctness.
ReSearch~\citep{chen2025research} and AutoRefine~\citep{shi2025autorefine} further improve this paradigm through richer trajectory formats and retrieval oriented training signals.
Search-E1 retains the standard outcome based GRPO formulation and focuses instead on whether the experience generated by this simple training loop can itself support further improvement.

\subsection{Beyond Outcome Based Reinforcement Learning}

A growing body of work augments outcome based reinforcement learning with additional supervision or training structure.
Thinker~\citep{xu2025thinker} distills decomposed search behavior from a substantially larger teacher, while StepSearch~\citep{wang2025stepsearch} uses externally generated intermediate supervision.
Other methods derive denser learning signals through process reward models or retrospective critics~\citep{zhang2025lessonsdevelopingprocessreward,lightman2023letsverifystepstep,zhang2025criticsearch}, policy based step estimators~\citep{wang2026igpo,feng2025gigpo}, or manually designed retrieval rewards~\citep{searchp1,hiprag,dynasearcher}.
A separate direction modifies exploration itself through tree structured sampling or multi stage optimization~\citep{ding2025treegrpo,hou2025treerl,jiang2025deepretrieval}.
These approaches differ substantially in mechanism, but share the goal of extracting supervision beyond a single trajectory level outcome.
Closest to our setting are methods that derive their additional signal from the policy itself rather than from an external system, either by conditioning a teacher on a hindsight summary of an entire rollout group and aligning the student at search query positions~\citep{ma2026sdsearch}, or by scoring each retrieval with an information gain estimate and modulating query token advantages~\citep{liang2026igsearch}.
Search-E1 differs from both in two respects.
Consolidation is a separate stage alternated with reinforcement learning rather than an auxiliary term inside a single update, and the privileged context is the search skeleton of one efficient successful trajectory rather than a group level summary or a scalar score.
Search-E1 studies a complementary question, namely whether such additional machinery is necessary when successful and unsuccessful behaviors already coexist in the policy's own rollout distribution.

\subsection{Self Distillation and Learning with Privileged Information}

Self distillation allows a model to provide supervision for itself, either through predictions from an earlier policy or through asymmetric conditioning between teacher and student~\citep{agarwal2024onpolicy,shenfeld2026continualsd}.
The latter is closely related to learning with privileged information~\citep{lopezpaz2016privileged}, where the teacher observes information that is unavailable to the student at inference time.
OPSD~\citep{zhao2026opsd} applies this principle to language model reasoning by conditioning the teacher on privileged reasoning information and distilling its token distribution into the standard policy.
Search-E1 builds on this mechanism but obtains the privileged information from the search skeleton of a sibling trajectory generated by the policy's own reinforcement learning process, so the teacher never observes retrieved evidence or the final answer.
More importantly, self distillation is not used as a terminal post training stage.
It is alternated with GRPO, so that reinforcement learning generates new experience and self distillation consolidates useful behaviors across successive rounds.
This repeated interaction between exploration and consolidation is the focus of Search-E1.

\clearpage
\section{Dataset Statistics}
\label{app:datasets}

Table~\ref{tab:datasets} summarizes the seven question answering benchmarks used in our experiments.
All datasets are obtained from the FlashRAG collection~\citep{FlashRAG}.
Following prior work~\citep{jin2025searchr1,shi2025autorefine}, we form the training corpus by combining the training splits of NQ and HotpotQA, which yields $169{,}615$ question answer pairs.
For evaluation we use the test split where one is available, namely NQ, TriviaQA, PopQA and Bamboogle, and the development split otherwise, namely HotpotQA, 2Wiki and MuSiQue.

\begin{table}[!htbp]
\centering
\caption{Statistics of the seven benchmarks used in this paper.}
\label{tab:datasets}
\setlength{\tabcolsep}{6pt}
\footnotesize
\begin{tabular}{lcccc}
\toprule
\multirow{2}{*}{\textbf{Dataset}} & \multicolumn{3}{c}{\textit{Split sizes}} & \multirow{2}{*}{\textbf{Eval split}} \\
\cmidrule(lr){2-4}
& Train & Dev & Test & \\
\midrule
NQ         & 79,168 & 8,757  & 3,610  & test \\
TriviaQA   & 78,785 & 8,837  & 11,313 & test \\
PopQA      & --     & --     & 14,267 & test \\
HotpotQA   & 90,447 & 7,405  & --     & dev  \\
2Wiki      & 15,000 & 12,576 & --     & dev  \\
MuSiQue    & 19,938 & 2,417  & --     & dev  \\
Bamboogle  & --     & --     & 125    & test \\
\bottomrule
\end{tabular}
\end{table}

\paragraph{Question pool across rounds.}
The self evolution loop maintains a question pool $\mathcal{D}_r$ that is initialized to the full training corpus.
A single GRPO round consumes far fewer instances than the corpus contains, so each round visits only a subset of the pool.
Rollout mining and the recycling rules of Equations~\ref{eq:opsd_data} and~\ref{eq:grpo_recycling} are applied to the questions visited in that round, and questions that were never visited remain in the pool unchanged.
Round wise pool statistics are reported in Appendix~\ref{app:data_recycling}.

\paragraph{Validation and calibration sets.}
Two auxiliary sets are used during training and never contribute to any reported result.
The validation set contains $1{,}000$ questions held out from the NQ and HotpotQA training splits, $500$ from each, and is excluded from the question pool and from every evaluation split.
All checkpoint selection, including the entropy aware handoff rule of Section~\ref{sec:evolvability}, is performed on this set, so no benchmark test or development split is consulted during training.
The calibration set contains $512$ additional held out questions and is kept fixed across all rounds.
It is used only to measure mean token entropy and rollout diversity, and no gradient is computed on it.

\clearpage
\section{Complete Training Configuration}
\label{app:hyperparams}

Table~\ref{tab:hyperparams} lists the original Search-E1 configuration; the additional experiments are described in Appendix~\ref{app:additional_experiments}.
Values not listed follow the default configuration of veRL~\citep{Sheng2025hybridflow}.
The GRPO block is shared with our internal GRPO controls; external baselines retain the configurations of their source papers or released code.
No per dataset tuning is performed, and the same configuration is used for the Base and Instruct variants at both scales.

\begin{table}[!htbp]
\centering
\caption{Complete configuration for Search-E1 training.}
\label{tab:hyperparams}
\setlength{\tabcolsep}{6pt}
\footnotesize
\begin{tabular}{ll}
\toprule
\textbf{Hyperparameter} & \textbf{Value} \\
\midrule
\multicolumn{2}{l}{\textit{Optimization}} \\
Framework & veRL \\
Hardware & 8 $\times$ H200 \\
Training batch size & 256 \\
Learning rate, GRPO and OPSD & $1 \times 10^{-6}$ \\
KL coefficient $\beta$ & 0.001 \\
Clip ratio $\epsilon$ & 0.2 \\
Max response length & 4096 tokens \\
\midrule
\multicolumn{2}{l}{\textit{Rollout and reward}} \\
Group size $G$ & 5 \\
Rollout temperature & 1.0 \\
Max search calls per rollout & 4 \\
Outcome reward & Exact Match \\
\midrule
\multicolumn{2}{l}{\textit{Retrieval}} \\
Retriever & E5-base-v2 \\
Corpus & Wikipedia (Dec. 2018) \\
Passages per query & 3 \\
\midrule
\multicolumn{2}{l}{\textit{Self evolution schedule}} \\
Cycles, 3B / 7B & 3 / 2 \\
GRPO steps, first round & 200 \\
GRPO steps, later rounds & 100 \\
Mining rollouts per visited question $K$ & 5 \\
Reference selection & fewest search calls, tie by skeleton length \\
Entropy retention threshold & 0.85 \\
\midrule
\multicolumn{2}{l}{\textit{OPSD}} \\
Divergence & forward KL \\
Pointwise clip $\tau_{\mathrm{clip}}$ & 10 \\
Alignment scope & $\mathcal{A}_{\tau}$ \\
Steps per round & 100 \\
Checkpoint interval & 20 steps \\
Privileged context max length & 512 tokens \\
Teacher & frozen copy of $\pi_r^{\mathrm{RL}}$ \\
\bottomrule
\end{tabular}
\end{table}

\paragraph{Schedule.}
The first GRPO round uses $200$ steps, matching the budget of the outcome reward baselines we compare against, and each later round uses $100$ steps.
Every OPSD round runs for $100$ steps and stores a checkpoint every $20$ steps, which yields six candidates including the initialization.
The checkpoint selected by the entropy aware handoff rule initializes the next GRPO round, and the remaining OPSD steps are discarded.
Appendix~\ref{app:budget} reports the full accounting of executed steps and wall clock time, including the discarded tail.

\clearpage
\section{Privileged Context Construction}
\label{app:privileged_context}

Section~\ref{sec:opsd} restricts the privileged context to the search skeleton of the reference trajectory.
This appendix specifies the resulting teacher input, isolates the contribution of each part of the reference that we discard, and reports the behavioral evidence behind the design.

\paragraph{Teacher input layout.}
Figure~\ref{fig:teacher_input} renders one teacher input as the model receives it.
The teacher sees three blocks in causal order, namely the task instruction with the question, the privileged skeleton, and the prefix of the student rollout up to the supervised position.
The student sees the same input with the skeleton block removed.
The difference between the two contexts at every supervised position is therefore exactly one block, and every supervised position has causal access to the whole skeleton.
At inference time the skeleton block is absent and the two views collapse into a single policy.

\begin{figure}[!htbp]
\centering
\footnotesize
\fbox{\parbox{0.94\linewidth}{
\textit{(1) Instruction and question, identical for student and teacher}\\[2pt]
\texttt{You are a helpful assistant. Answer the question by reasoning}\\
\texttt{and searching when needed. Question: What is the date of death}\\
\texttt{of Alexander of Masovia's father?}\\[6pt]
\colorbox{yellow!25}{\parbox{0.97\linewidth}{
\textit{(2) Privileged skeleton, teacher only}\\[2pt]
\texttt{[Reference search plan]}\\
\texttt{<search> Alexander of Masovia father </search> ->}\\
\texttt{<search> Siemowit IV Duke of Masovia date of death </search>}
}}\\[6pt]
\textit{(3) Student rollout prefix, identical for student and teacher}\\[2pt]
\texttt{<think> First identify the father, then look up his death}\\
\texttt{date. </think> <search> Alexander of Masovia father </search>}\\
\texttt{<information> ... </information> <think> Now I need the}\\
\texttt{death date of Siemowit IV. </think> <search>}\\[6pt]
\textit{(4) Supervised positions, forward KL applied over $\mathcal{A}_{\tau^{\mathrm{stu}}}$}
}}
\caption{
Teacher input layout for one supervised position.
Block (2) contains only the \texttt{<search>} spans of the reference trajectory, in their original order.
Intermediate reasoning, retrieved passages and the final answer of the reference are removed by the masking operator $\mathcal{M}$.
The student input is identical with block (2) deleted, so the difference between the two contexts is the shaded block.
}
\label{fig:teacher_input}
\end{figure}

\paragraph{What the teacher should and should not see.}
We vary the content of the privileged block while keeping the rest of the pipeline fixed, and evaluate each variant under the full three cycle schedule with Qwen2.5-3B-Instruct.
Beyond Exact Match we report two behavioral quantities measured on the evaluation benchmarks, namely the average number of search calls per rollout and the search quality, defined as the fraction of retrieval calls whose returned passages contain the gold answer.
We also report the average length of the privileged block in tokens.
Table~\ref{tab:privileged_context} reports the results.

\begin{table}[!htbp]
\centering
\caption{
Privileged context construction on Qwen2.5-3B-Instruct.
Single and Multi denote the average Exact Match over the three single hop and four multi hop benchmarks.
Calls is the average number of search calls per rollout and Quality is the fraction of retrieval calls returning the gold answer.
Length is the average size of the privileged block in tokens.
}
\label{tab:privileged_context}
\setlength{\tabcolsep}{5pt}
\footnotesize
\begin{tabular}{lcccccc}
\toprule
\textbf{Configuration} & \textbf{Single} & \textbf{Multi} & \textbf{Avg.} & \textbf{Calls} & \textbf{Quality} & \textbf{Length} \\
\midrule
Search skeleton (Search-E1) & 0.520 & 0.380 & \textbf{0.440} & 2.12 & 54.3 & 41 \\
\midrule
Skeleton with shuffled query order & 0.515 & 0.368 & 0.431 & 2.09 & 51.8 & 41 \\
Skeleton with final answer appended & 0.509 & 0.362 & 0.425 & 1.91 & 48.7 & 47 \\
Skeleton with retrieved passages & 0.512 & 0.365 & 0.428 & 1.74 & 50.1 & 1183 \\
Full trajectory & 0.506 & 0.356 & 0.420 & 1.38 & 46.2 & 1930 \\
No privileged context & 0.507 & 0.365 & 0.426 & 2.05 & 52.0 & 0 \\
\bottomrule
\end{tabular}
\end{table}

Three observations follow.
First, exposing the answer is harmful rather than merely redundant.
Appending the final answer of the reference costs $1.5$ points, and passing the full trajectory costs $2.0$ points, which places it below the $0.426$ obtained without any distillation at all.
The behavioral columns explain why.
As more of the reference becomes visible, the teacher can predict the gold answer from its own context without issuing a query, so its distribution at search positions concentrates on continuations that terminate retrieval early.
Search calls per rollout fall from $2.12$ to $1.38$ and search quality falls by $8.1$ points.
The student inherits this behavior and answers on weaker evidence.
A teacher that already knows the answer is therefore not a stronger teacher, and the value of the privileged context lies in the search plan rather than in the outcome.

Second, the ordering of the skeleton carries information.
Shuffling the queries of the reference while keeping the same set costs $0.9$ points, with the loss concentrated on the multi hop benchmarks, where the second query is typically conditioned on the entity resolved by the first.
The skeleton is thus read by the teacher as a plan rather than as a bag of keywords.

Third, the skeleton is also the cheapest option.
It averages $41$ tokens, against $1{,}930$ tokens for the full trajectory, because retrieved passages dominate the length of a search augmented trajectory.
Restricting the privileged context to the skeleton therefore removes the leakage, improves accuracy, and keeps the teacher forward pass close to the cost of the student pass.

The last row is a degenerate control in which the teacher and the student receive identical contexts.
The forward KL term is then zero by construction and training reduces to GRPO with the same recycling schedule, which recovers the GRPO only entry of Table~\ref{tab:ablation} and confirms that the gain is attributable to the asymmetry of the conditioning rather than to the additional optimization stage.

\clearpage
\section{Training Budget and Compute Accounting}
\label{app:budget}

Search-E1 executes more optimizer steps than the single stage baselines it is compared against, so a natural concern is whether the reported gains reflect the alternation of exploration and consolidation or simply a larger training budget.
This appendix reports the full cost of a self evolution run and compares Search-E1 against a control matched on executed optimizer steps.

\paragraph{Cost of one self evolution run.}
On 8 $\times$ H200 with Qwen2.5-3B and a batch size of $256$, one GRPO step takes $60$ seconds and one OPSD step takes $66$ seconds.
Rollout generation accounts for $44$ seconds of both, so the components specific to OPSD, namely the teacher forward pass over the privileged context and the token level divergence, add roughly one tenth of a step.
Table~\ref{tab:budget} reports the resulting breakdown.
The accounting includes the rollout mining stages and the checkpoint evaluations required by the handoff rule, and it charges each OPSD round for all $100$ executed steps even though the tail after the selected checkpoint is discarded.

\begin{table}[!htbp]
\centering
\caption{
Wall clock breakdown of one complete Search-E1 run on Qwen2.5-3B, 8 $\times$ H200, batch size 256.
Steps counts optimizer steps.
Mining and checkpoint selection perform generation and evaluation only.
}
\label{tab:budget}
\setlength{\tabcolsep}{8pt}
\footnotesize
\begin{tabular}{lcc}
\toprule
\textbf{Stage} & \textbf{Steps} & \textbf{Wall clock (h)} \\
\midrule
GRPO round 1 & 200 & 3.33 \\
Rollout mining 1 & -- & 2.44 \\
OPSD round 1 & 100 & 1.83 \\
GRPO round 2 & 100 & 1.67 \\
Rollout mining 2 & -- & 1.22 \\
OPSD round 2 & 100 & 1.83 \\
GRPO round 3 & 100 & 1.67 \\
Rollout mining 3 & -- & 1.22 \\
OPSD round 3 & 100 & 1.83 \\
Checkpoint selection & -- & 0.30 \\
\midrule
Total & 700 & 17.36 \\
\bottomrule
\end{tabular}
\end{table}

The complete run takes $17.4$ hours, against $3.5$ hours for a $200$ step outcome reward baseline trained under the same hardware, batch size, rollout configuration and retriever.
Two properties of this cost are worth noting.
It is entirely contained within the standard reinforcement learning loop, since no external model is queried and no annotation phase is performed, and it leaves inference unchanged, because the privileged context and the teacher forward pass are removed at test time.
Mining accounts for $28$ percent of the total, and its cost is proportional to the number of questions visited in the preceding round rather than to the size of the corpus.

\begin{figure}[!htbp]
\centering
\includegraphics[width=0.55\linewidth]{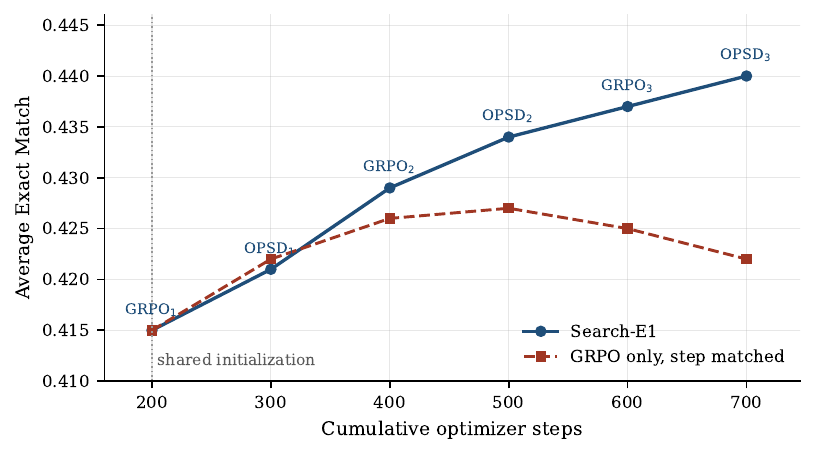}
\caption{
Average Exact Match against cumulative optimizer steps on Qwen2.5-3B-Instruct.
Both runs share the first $200$ GRPO steps and the same recycling schedule.
The Search-E1 markers report the checkpoint selected at the end of each stage.
The control continues with GRPO alone for the same total number of steps.
}
\label{fig:budget_curve}
\end{figure}

\paragraph{Step matched comparison.}
To separate the effect of alternation from the effect of budget, we train a control that runs GRPO alone for $700$ optimizer steps, which is the total number of steps executed by Search-E1, under the identical recycling schedule and starting from the same round one checkpoint.
This control costs $11.8$ hours, so Search-E1 spends $1.47$ times its wall clock.
Figure~\ref{fig:budget_curve} plots average Exact Match against cumulative optimizer steps for both runs.

The two runs coincide at step $200$ by construction and then diverge.
The control improves to $0.426$ at step $400$, peaks at $0.427$ at step $500$, and declines to $0.422$ by step $700$, whereas Search-E1 reaches $0.440$.
The decline of the control is not caused by a loss of exploration capacity, since its mean token entropy on the calibration set still retains $0.93$ of its step $200$ value at step $700$, well above the level at which the handoff rule intervenes.
Additional outcome reward optimization on an already recycled pool therefore stops converting rollouts into improvement, while the same rollouts still carry contrasts that consolidation can extract.

\paragraph{Comparison with externally supervised pipelines.}
The methods that Search-E1 outperforms in Tables~\ref{tab:main_3b} and~\ref{tab:main_7b} carry costs that their reported training time does not include.
Thinker~\citep{xu2025thinker} uses teacher-generated training data; the Thinker rows in our main tables are its SFT results reported by SD-Search. StepSearch~\citep{wang2025stepsearch} uses intermediate sub-question supervision. These are additional supervision sources, whose costs are not measured by our step-matched GRPO comparison.
Search-E1 instead uses self generated trajectories and reference answers; the measured cost comparison in this appendix is against our own GRPO control.

\clearpage
\section{Full Results Across Self Evolution Rounds}
\label{app:full_results}

Figure~\ref{fig:self_evolution} in the main text reports only the seven benchmark average after each stage.
This appendix gives the complete per benchmark results for every stage of every backbone.
GRPO$_r$ denotes the policy at the end of the $r$ th exploration round.
OPSD$_r$ denotes the checkpoint selected by the entropy aware handoff rule at the end of the $r$ th consolidation round, which is also the checkpoint that initializes the following GRPO round.
The last row of each table reproduces the corresponding entry of Table~\ref{tab:main_3b} or Table~\ref{tab:main_7b}.

\begin{table}[!htbp]
\centering
\caption{
Exact Match after every stage of Search-E1 training with Qwen2.5-3B-Base.
GRPO$_r$ denotes the policy at the end of the $r$ th exploration round and OPSD$_r$ the checkpoint selected by the entropy aware handoff rule at the end of the $r$ th consolidation round.
The final row reproduces the corresponding entry of the main results table.
}
\label{tab:rounds_3b_base}
\setlength{\tabcolsep}{4pt}
\footnotesize
\begin{tabular}{l ccc cccc c}
\toprule
& \multicolumn{3}{c}{\textit{Single-Hop}} & \multicolumn{4}{c}{\textit{Multi-Hop}} & \\
\cmidrule(lr){2-4} \cmidrule(lr){5-8}
\textbf{Stage} & NQ & TriviaQA & PopQA & HotpotQA & 2Wiki & MuSiQue & Bamboogle & \textbf{Avg.} \\
\midrule
GRPO$_1$ & 0.461 & 0.623 & 0.460 & 0.405 & 0.400 & 0.165 & 0.377 & 0.413 \\
OPSD$_1$ & 0.466 & 0.628 & 0.465 & 0.411 & 0.406 & 0.171 & 0.386 & 0.419 \\
GRPO$_2$ & 0.471 & 0.633 & 0.470 & 0.420 & 0.417 & 0.179 & 0.399 & 0.427 \\
OPSD$_2$ & 0.475 & 0.637 & 0.474 & 0.425 & 0.422 & 0.184 & 0.407 & 0.432 \\
GRPO$_3$ & 0.477 & 0.639 & 0.476 & 0.429 & 0.425 & 0.187 & 0.412 & 0.435 \\
\midrule
\textbf{OPSD$_3$} & \textbf{0.479} & \textbf{0.641} & \textbf{0.478} & \textbf{0.432} & \textbf{0.429} & \textbf{0.190} & \textbf{0.416} & \textbf{0.438} \\
\bottomrule
\end{tabular}
\end{table}

\begin{table}[!htbp]
\centering
\caption{
Exact Match after every stage of Search-E1 training with Qwen2.5-3B-Instruct.
GRPO$_r$ denotes the policy at the end of the $r$ th exploration round and OPSD$_r$ the checkpoint selected by the entropy aware handoff rule at the end of the $r$ th consolidation round.
The final row reproduces the corresponding entry of the main results table.
}
\label{tab:rounds_3b_instruct}
\setlength{\tabcolsep}{4pt}
\footnotesize
\begin{tabular}{l ccc cccc c}
\toprule
& \multicolumn{3}{c}{\textit{Single-Hop}} & \multicolumn{4}{c}{\textit{Multi-Hop}} & \\
\cmidrule(lr){2-4} \cmidrule(lr){5-8}
\textbf{Stage} & NQ & TriviaQA & PopQA & HotpotQA & 2Wiki & MuSiQue & Bamboogle & \textbf{Avg.} \\
\midrule
GRPO$_1$ & 0.456 & 0.609 & 0.444 & 0.399 & 0.406 & 0.168 & 0.423 & 0.415 \\
OPSD$_1$ & 0.461 & 0.613 & 0.448 & 0.406 & 0.412 & 0.174 & 0.433 & 0.421 \\
GRPO$_2$ & 0.466 & 0.618 & 0.453 & 0.415 & 0.423 & 0.182 & 0.446 & 0.429 \\
OPSD$_2$ & 0.470 & 0.622 & 0.457 & 0.420 & 0.429 & 0.187 & 0.453 & 0.434 \\
GRPO$_3$ & 0.472 & 0.624 & 0.459 & 0.424 & 0.432 & 0.190 & 0.458 & 0.437 \\
\midrule
\textbf{OPSD$_3$} & \textbf{0.474} & \textbf{0.626} & \textbf{0.461} & \textbf{0.427} & \textbf{0.436} & \textbf{0.193} & \textbf{0.464} & \textbf{0.440} \\
\bottomrule
\end{tabular}
\end{table}

\begin{table}[!htbp]
\centering
\caption{
Exact Match after every stage of Search-E1 training with Qwen2.5-7B-Base.
GRPO$_r$ denotes the policy at the end of the $r$ th exploration round and OPSD$_r$ the checkpoint selected by the entropy aware handoff rule at the end of the $r$ th consolidation round.
The final row reproduces the corresponding entry of the main results table.
}
\label{tab:rounds_7b_base}
\setlength{\tabcolsep}{4pt}
\footnotesize
\begin{tabular}{l ccc cccc c}
\toprule
& \multicolumn{3}{c}{\textit{Single-Hop}} & \multicolumn{4}{c}{\textit{Multi-Hop}} & \\
\cmidrule(lr){2-4} \cmidrule(lr){5-8}
\textbf{Stage} & NQ & TriviaQA & PopQA & HotpotQA & 2Wiki & MuSiQue & Bamboogle & \textbf{Avg.} \\
\midrule
GRPO$_1$ & 0.483 & 0.663 & 0.485 & 0.440 & 0.412 & 0.196 & 0.478 & 0.451 \\
OPSD$_1$ & 0.488 & 0.668 & 0.490 & 0.449 & 0.422 & 0.204 & 0.492 & 0.459 \\
GRPO$_2$ & 0.497 & 0.677 & 0.499 & 0.462 & 0.436 & 0.216 & 0.510 & 0.471 \\
\midrule
\textbf{OPSD$_2$} & \textbf{0.503} & \textbf{0.683} & \textbf{0.505} & \textbf{0.472} & \textbf{0.447} & \textbf{0.225} & \textbf{0.525} & \textbf{0.480} \\
\bottomrule
\end{tabular}
\end{table}

\begin{table}[t]
\centering
\caption{
Exact Match after every stage of Search-E1 training with Qwen2.5-7B-Instruct.
GRPO$_r$ denotes the policy at the end of the $r$ th exploration round and OPSD$_r$ the checkpoint selected by the entropy aware handoff rule at the end of the $r$ th consolidation round.
The final row reproduces the corresponding entry of the main results table.
}
\label{tab:rounds_7b_instruct}
\setlength{\tabcolsep}{4pt}
\footnotesize
\begin{tabular}{l ccc cccc c}
\toprule
& \multicolumn{3}{c}{\textit{Single-Hop}} & \multicolumn{4}{c}{\textit{Multi-Hop}} & \\
\cmidrule(lr){2-4} \cmidrule(lr){5-8}
\textbf{Stage} & NQ & TriviaQA & PopQA & HotpotQA & 2Wiki & MuSiQue & Bamboogle & \textbf{Avg.} \\
\midrule
GRPO$_1$ & 0.464 & 0.671 & 0.474 & 0.433 & 0.416 & 0.203 & 0.482 & 0.449 \\
OPSD$_1$ & 0.470 & 0.677 & 0.480 & 0.443 & 0.427 & 0.212 & 0.497 & 0.458 \\
GRPO$_2$ & 0.480 & 0.687 & 0.490 & 0.458 & 0.444 & 0.226 & 0.519 & 0.472 \\
\midrule
\textbf{OPSD$_2$} & \textbf{0.487} & \textbf{0.694} & \textbf{0.497} & \textbf{0.469} & \textbf{0.455} & \textbf{0.236} & \textbf{0.534} & \textbf{0.482} \\
\bottomrule
\end{tabular}
\end{table}

Three patterns hold across all four backbones.
Every consolidation stage improves on the exploration stage that precedes it, and every exploration stage improves on the consolidation stage that precedes it, so neither stage merely undoes the other.
The per stage gains are larger on the multi hop benchmarks throughout, which is consistent with the interpretation that a privileged search plan is most informative when the question requires several dependent queries.
The gains contract from round to round, and the third cycle at the 3B scale contributes less than half of what the first contributes.
We did not observe a round at which performance decreased, but we also did not extend the schedule beyond the point at which the marginal gain fell below the seed level variation reported in Appendix~\ref{app:seeds}.

\clearpage
\section{Alternatives to Consolidation}
\label{app:consolidation}

Search-E1 consumes its own successful rollouts through on policy self distillation.
Two simpler alternatives exist.
The experience can be consumed by supervised fine tuning on the mined trajectories, and the consolidation stage can be applied once instead of being alternated with exploration.
Table~\ref{tab:consolidation} compares these under the full three cycle schedule with Qwen2.5-3B-Instruct.

For the supervised variants, each OPSD round is replaced by $100$ steps of supervised fine tuning at the same learning rate on the same mined references, with the loss taken over policy generated positions only, and the resulting checkpoints are passed to the next round by the same entropy aware rule.
The two supervised variants differ in the target sequence, namely the complete successful trajectory or its search skeleton alone.
For the single stage variants, the entire consolidation budget of $300$ steps is spent at one point in the schedule, either after the first GRPO round or after all three.
The last column reports the entropy retention measured on the calibration set at the end of the consolidation stage, before checkpoint selection.

\begin{table}[!htbp]
\centering
\caption{
Alternatives to on policy self distillation on Qwen2.5-3B-Instruct.
Single and Multi denote the average Exact Match over the three single hop and four multi hop benchmarks.
Entropy is the entropy retention on the calibration set at the end of the consolidation stage, before checkpoint selection.
}
\label{tab:consolidation}
\setlength{\tabcolsep}{6pt}
\footnotesize
\begin{tabular}{lcccc}
\toprule
\textbf{Configuration} & \textbf{Single} & \textbf{Multi} & \textbf{Avg.} & \textbf{Entropy} \\
\midrule
OPSD, alternating (Search-E1) & 0.520 & 0.380 & \textbf{0.440} & 0.69 \\
\midrule
Distillation once after the first round & 0.513 & 0.368 & 0.430 & 0.69 \\
Distillation once after all GRPO rounds & 0.514 & 0.370 & 0.432 & 0.68 \\
SFT on mined trajectories, alternating & 0.512 & 0.365 & 0.428 & 0.52 \\
SFT on mined skeletons, alternating & 0.509 & 0.361 & 0.424 & 0.58 \\
No consolidation (GRPO only) & 0.507 & 0.365 & 0.426 & -- \\
\bottomrule
\end{tabular}
\end{table}

Supervised fine tuning recovers only a small part of the benefit.
Fitting the complete mined trajectory reaches $0.428$ and fitting the skeleton alone reaches $0.424$, against $0.440$ for distillation.
Two differences account for the gap.
The supervised objective is evaluated on the reference sequence, so it supplies no signal at the positions the student actually visits when it departs from that sequence, whereas the distillation objective is evaluated along the student's own rollout.
The supervised objective also drives the policy toward a single realization rather than toward a distribution, and the entropy column shows the consequence.
Supervised fine tuning removes roughly half of the policy entropy within one hundred steps, against thirty one percent for distillation, so under the same handoff constraint it is forced to stop far earlier and consolidates correspondingly less.
This is the same mechanism examined in Appendix~\ref{app:entropy_diversity}, observed here as a property of the objective rather than of the number of steps.

Applying distillation once rather than alternating is also worse.
Spending the entire consolidation budget after the first GRPO round reaches $0.430$, and spending it after all three rounds reaches $0.432$.
Neither approaches the alternating schedule.
When consolidation happens only at the beginning, the behaviors it absorbs are those of an early policy and become stale as exploration continues.
When it happens only at the end, the exploration rounds never benefit from the behaviors that earlier rounds discovered, so each round rediscovers ground that a consolidated policy would already hold.
The value of consolidation therefore lies in its position within the loop and not only in the objective it optimizes.

\clearpage
\section{Question Recycling and Pool Composition}
\label{app:data_recycling}

Section~\ref{sec:selfevolution} retains for the next GRPO round every visited question that is not uniformly solved, so all failure questions and mixed outcome questions are carried forward while uniformly solved ones are dropped.
This appendix reports how the rollout pool is composed across rounds and compares this rule against three alternatives.

\paragraph{Pool composition across rounds.}
After each GRPO round we sample $K=5$ trajectories for every question visited during that round and partition the questions by the number of correct trajectories $c_r(q)$.
Table~\ref{tab:pool_stats} reports the resulting composition on Qwen2.5-3B-Instruct.

\begin{table}[!htbp]
\centering
\caption{
Rollout pool composition after each GRPO round on Qwen2.5-3B-Instruct.
Percentages are taken over the questions visited in that round.
References counts the questions that supply a privileged reference for the following OPSD round.
Pool size is the number of questions remaining after uniformly solved ones are removed.
}
\label{tab:pool_stats}
\setlength{\tabcolsep}{6pt}
\footnotesize
\begin{tabular}{lcccccc}
\toprule
\textbf{Round} & \textbf{Visited} & $c_r{=}0$ (\%) & $0{<}c_r{<}K$ (\%) & $c_r{=}K$ (\%) & \textbf{References} & \textbf{Pool size} \\
\midrule
1 & 51,200 & 38.0 & 44.0 & 18.0 & 31,744 & 160,399 \\
2 & 25,600 & 31.0 & 46.0 & 23.0 & 17,664 & 154,511 \\
3 & 25,600 & 27.0 & 47.0 & 26.0 & 18,688 & 147,855 \\
\bottomrule
\end{tabular}
\end{table}

\begin{figure}[!htbp]
\centering
\includegraphics[width=0.55\linewidth]{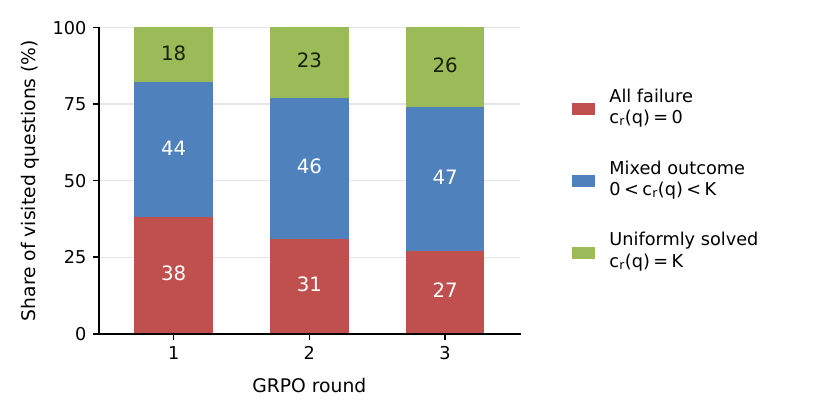}
\caption{
Composition of the rollout pool after each GRPO round on Qwen2.5-3B-Instruct.
The share of uniformly solved questions grows across rounds while the share of all failure questions shrinks, so the two groups that the recycling rule retains change in opposite directions as training proceeds.
}
\label{fig:pool_composition}
\end{figure}

Two trends are visible, and Figure~\ref{fig:pool_composition} shows the same composition.
The fraction of uniformly solved questions grows from $18.0$ to $26.0$ percent as the policy improves, so an increasing share of every round would otherwise be spent on questions from which GRPO can extract no advantage signal, since a group in which every trajectory succeeds yields a zero normalized advantage exactly as one in which every trajectory fails.
At the same time the fraction of all failure questions falls from $38.0$ to $27.0$ percent, which means the questions that supply no privileged reference are also the ones that shrink fastest.
The mixed outcome group, which contributes to both stages, remains close to half of every round throughout training.

\paragraph{Which questions should be retained.}
We compare four recycling rules under the full three cycle schedule, keeping everything else fixed.
The alternatives retain only mixed outcome questions, retain only all failure questions, or retain every visited question regardless of outcome.
Table~\ref{tab:recycling} reports the results.

\begin{table}[t]
\centering
\caption{
Question recycling strategies on Qwen2.5-3B-Instruct.
Single and Multi denote the average Exact Match over the three single hop and four multi hop benchmarks.
}
\label{tab:recycling}
\setlength{\tabcolsep}{6pt}
\footnotesize
\begin{tabular}{lccc}
\toprule
\textbf{Retained for the next round} & \textbf{Single} & \textbf{Multi} & \textbf{Avg.} \\
\midrule
Failure and mixed outcome (Search-E1) & 0.520 & 0.380 & \textbf{0.440} \\
Mixed outcome only & 0.512 & 0.371 & 0.431 \\
All failure only & 0.505 & 0.367 & 0.426 \\
Retain every visited question & 0.510 & 0.374 & 0.432 \\
\bottomrule
\end{tabular}
\end{table}

Retaining every visited question costs $0.8$ points.
Uniformly solved questions produce a zero advantage under group normalization, so they consume rollout budget without contributing gradient, and their share grows with every round.
Removing them concentrates later exploration on the questions where search behavior can still change.

Discarding either of the two informative groups is worse than keeping both.
Retaining mixed outcome questions alone costs $0.9$ points, and retaining all failure questions alone costs $1.4$ points.
The two groups play different roles.
Mixed outcome questions are the only source of privileged references among the questions that also remain useful for exploration, so dropping them starves the consolidation stage.
All failure questions supply no reference in the current round, yet they are the questions on which the policy has the most room to improve, and Table~\ref{tab:pool_stats} shows that a substantial part of them converts into mixed outcome questions in later rounds.
Keeping both groups is therefore not a compromise between two criteria but a consequence of the two stages consuming different parts of the same pool.

\clearpage
\section{Privileged Reference Selection}
\label{app:reference_selection}

By default Search-E1 selects, among the successful rollouts of a question, the one with the fewest search calls, breaking ties by the total length of its skeleton.
Table~\ref{tab:reference_selection} compares this rule against five alternatives under the full three cycle schedule with Qwen2.5-3B-Instruct.
The last column reports the average number of search calls in the selected reference, which is the quantity the selection rule most directly controls.

\begin{table}[!htbp]
\centering
\caption{
Privileged reference selection on Qwen2.5-3B-Instruct.
Reference calls is the average number of search calls in the selected reference trajectory.
}
\label{tab:reference_selection}
\setlength{\tabcolsep}{6pt}
\footnotesize
\begin{tabular}{lcccc}
\toprule
\textbf{Selection rule} & \textbf{Single} & \textbf{Multi} & \textbf{Avg.} & \textbf{Reference calls} \\
\midrule
Fewest search calls (Search-E1) & 0.520 & 0.380 & \textbf{0.440} & 1.68 \\
Shortest skeleton in tokens & 0.518 & 0.377 & 0.437 & 1.79 \\
Random successful trajectory & 0.516 & 0.374 & 0.435 & 2.24 \\
Longest successful trajectory & 0.510 & 0.372 & 0.431 & 3.12 \\
Most search calls & 0.508 & 0.370 & 0.429 & 3.31 \\
Random rollout regardless of outcome & 0.503 & 0.358 & 0.420 & 2.41 \\
\bottomrule
\end{tabular}
\end{table}

Performance decreases monotonically as the selected reference becomes less efficient, from $0.440$ at $1.68$ average calls to $0.429$ at $3.31$.
A successful trajectory that reaches the answer after several redundant queries still demonstrates a path to the answer, but the redundant steps are part of the plan the teacher conditions on, so the teacher rewards them at the student's search positions as readily as the productive ones.
Selecting by skeleton length rather than by call count performs almost as well at $0.437$, which indicates that the useful notion of efficiency is the number of retrieval decisions rather than verbosity, although the two are correlated.

The final row removes the correctness requirement entirely and selects a uniformly random rollout as the reference.
This reaches $0.420$, below the $0.426$ obtained without any consolidation, and is the only variant in this table that is actively harmful.
Conditioning on an unsuccessful plan makes the teacher a worse predictor than the student at exactly the positions the objective targets, so the distillation term transfers failure modes rather than removing them.
Correctness is therefore not a convenience for filtering but the property that makes the asymmetry between teacher and student informative, which also explains why the method inherits the requirement for automatically checkable answers.

\clearpage
\section{Distillation Objective and Alignment Scope}
\label{app:objective}

The consolidation stage makes two choices, namely the divergence between the teacher and student distributions and the set of positions over which it is summed.
We vary each while keeping everything else fixed, under the full three cycle schedule with Qwen2.5-3B-Instruct.
Table~\ref{tab:objective} reports both the final Exact Match and the entropy retention at the end of the consolidation round, because the two are not ordered in the same way and the second determines how much of the round the handoff rule can use.

\begin{table}[!htbp]
\centering
\caption{
Divergence and alignment scope on Qwen2.5-3B-Instruct.
Entropy is the entropy retention on the calibration set at the end of the second consolidation round, before checkpoint selection.
}
\label{tab:objective}
\setlength{\tabcolsep}{6pt}
\footnotesize
\begin{tabular}{lcccc}
\toprule
\textbf{Configuration} & \textbf{Single} & \textbf{Multi} & \textbf{Avg.} & \textbf{Entropy} \\
\midrule
\multicolumn{5}{l}{\textit{Divergence, scope fixed to $\mathcal{A}_\tau$}} \\
\midrule
Forward KL with pointwise clipping (Search-E1) & 0.520 & 0.380 & \textbf{0.440} & 0.69 \\
Forward KL without clipping & 0.514 & 0.372 & 0.433 & 0.66 \\
Jensen-Shannon divergence & 0.517 & 0.375 & 0.436 & 0.74 \\
Reverse KL & 0.512 & 0.365 & 0.428 & 0.41 \\
Mean squared error on logits & 0.506 & 0.358 & 0.421 & 0.58 \\
\midrule
\multicolumn{5}{l}{\textit{Alignment scope, divergence fixed to clipped forward KL}} \\
\midrule
All policy generated positions (Search-E1) & 0.520 & 0.380 & \textbf{0.440} & 0.69 \\
Reasoning and search positions & 0.518 & 0.377 & 0.437 & 0.71 \\
Search query positions only & 0.513 & 0.375 & 0.434 & 0.79 \\
Answer positions only & 0.509 & 0.362 & 0.425 & 0.83 \\
\bottomrule
\end{tabular}
\end{table}

\paragraph{Divergence.}
Reverse KL is the weakest of the three divergence choices at $0.428$, and it is also the one that sharpens the policy most, retaining only $0.41$ of the entropy within one hundred steps.
Its mode seeking behavior collapses the student onto the teacher's most likely continuation, which suppresses precisely the variation among sibling rollouts that the next exploration round needs.
This connects the objective to the evolvability analysis of Section~\ref{sec:evolvability}, since a divergence that sharpens faster leaves the handoff rule a smaller usable portion of each round.
The Jensen-Shannon divergence is the closest competitor at $0.436$ and preserves the most entropy, but it is bounded and symmetric, so at the many positions where teacher and student already agree it attenuates an already small signal, and the resulting update is slower without being better targeted.
Removing the pointwise clip from the forward KL costs $0.7$ points and produces visibly noisier training curves, because a small number of vocabulary entries with very large teacher probability ratios dominate individual updates.
Mean squared error on the raw logits performs worst, as it treats every coordinate as equally important and ignores the simplex structure that the divergences respect.

\paragraph{Alignment scope.}
Narrowing the scope to the search query positions costs $0.6$ points.
This is a smaller effect than any of the divergence changes, which is expected since the query positions carry most of the behavioral difference the privileged plan induces, but it is consistent.
The privileged context is a plan rather than an answer, so it also changes how the policy reasons between retrievals, and aligning the reasoning positions transfers part of that.
Restricting the objective to the answer positions alone is the weakest configuration at $0.425$, close to no consolidation, confirming that the transferred behavior concerns how evidence is gathered rather than how the final string is produced.

\clearpage
\section{Entropy and Rollout Diversity}
\label{app:entropy_diversity}

Section~\ref{sec:evolvability} uses mean token entropy as the criterion for passing a checkpoint to the next exploration round.
Entropy is a per position quantity, so it is not obvious that it tracks the trajectory level diversity that GRPO actually consumes.
This appendix measures both on the same checkpoints of the second consolidation round of Qwen2.5-3B-Instruct.

For each checkpoint we sample $K=5$ rollouts for every question of the calibration set and report four quantities.
Distinct trajectories counts how many of the five rollouts are pairwise different after normalizing whitespace.
Distinct first queries counts how many different opening queries appear among the five.
Pairwise distance is the mean normalized edit distance between the search skeletons of two sibling rollouts.
Reference coverage is the fraction of questions for which at least one of the five rollouts is correct, which is the fraction that would supply a privileged reference in the next mining stage.

\begin{table}[!htbp]
\centering
\caption{
Entropy and rollout diversity over the second consolidation round of Qwen2.5-3B-Instruct.
All quantities are measured on the fixed calibration set with $K=5$ rollouts per question.
Entropy is expressed relative to the value at the beginning of the round.
}
\label{tab:diversity}
\setlength{\tabcolsep}{6pt}
\footnotesize
\begin{tabular}{cccccc}
\toprule
\textbf{OPSD step} & \textbf{Entropy} & \textbf{Distinct traj.} & \textbf{Distinct first queries} & \textbf{Pairwise dist.} & \textbf{Coverage} \\
\midrule
0 & 1.00 & 4.31 & 3.72 & 0.62 & 0.69 \\
20 & 0.95 & 4.18 & 3.61 & 0.60 & 0.70 \\
40 & 0.91 & 4.02 & 3.45 & 0.57 & 0.71 \\
60 & 0.86 & 3.84 & 3.28 & 0.54 & 0.71 \\
80 & 0.78 & 3.46 & 2.90 & 0.47 & 0.70 \\
100 & 0.69 & 3.02 & 2.44 & 0.39 & 0.68 \\
\bottomrule
\end{tabular}
\end{table}

\begin{figure}[!htbp]
\centering
\includegraphics[width=0.55\linewidth]{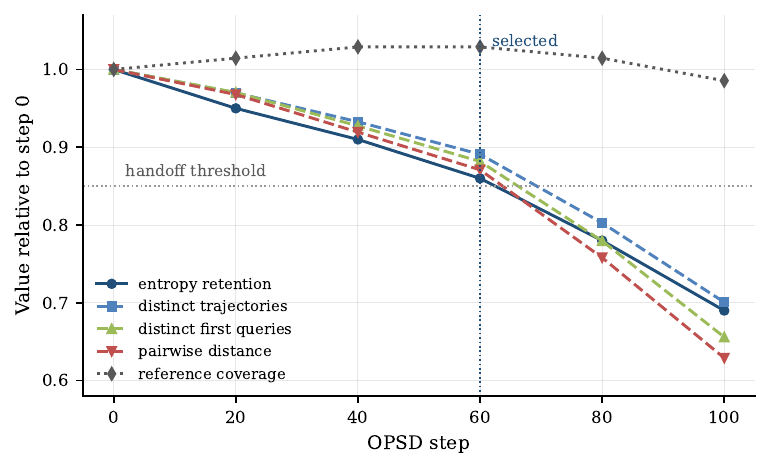}
\caption{
Entropy retention and rollout diversity over the second consolidation round of Qwen2.5-3B-Instruct.
Diversity quantities are normalized by their value at the beginning of the round.
The three diversity measures track entropy closely, while reference coverage stays nearly flat until entropy falls below the handoff threshold.
}
\label{fig:diversity}
\end{figure}

The three diversity measures track entropy closely and decline faster than linearly in the second half of the round.
Between step $60$ and step $100$ the entropy retention falls from $0.86$ to $0.69$, while the number of distinct trajectories among five falls from $3.84$ to $3.02$ and the number of distinct opening queries falls from $3.28$ to $2.44$.
At that point roughly half of the sampled rollouts for a question begin with the same query, so a group of five carries substantially less information than its size suggests.
Since the group normalized advantage of Equation~\ref{eq:grpo_adv} is defined by the spread of outcomes within a group, this directly reduces the signal available to the next exploration round.

Reference coverage behaves differently.
It rises slightly through the first half of the round as consolidation makes the policy more reliable, peaks at $0.71$, and then falls back to $0.68$ once the collapse in diversity outweighs the improvement in per rollout accuracy.
The checkpoint selected by the handoff rule sits at the peak of this quantity, which is a useful coincidence rather than a criterion we optimize, since coverage can only be measured after the following mining stage while entropy is available during the round itself.

\clearpage
\section{Entropy Aware Handoff Threshold}
\label{app:handoff}

The handoff rule keeps checkpoints whose entropy retention is at least $0.85$ and selects the best of these by validation Exact Match.
Table~\ref{tab:handoff} varies this threshold.
For each value we report the checkpoint the rule selects in the second consolidation round, its immediate validation Exact Match, the Exact Match reached by the exploration round initialized from it, and the average Exact Match at the end of the complete schedule.

\begin{table}[!htbp]
\centering
\caption{
Entropy retention threshold on Qwen2.5-3B-Instruct.
Selected is the OPSD step chosen by the rule in the second consolidation round.
Immediate and Next GRPO are validation Exact Match before and after the following exploration round.
Final is the seven benchmark average at the end of the complete schedule.
}
\label{tab:handoff}
\setlength{\tabcolsep}{6pt}
\footnotesize
\begin{tabular}{cccccc}
\toprule
\textbf{Threshold} & \textbf{Selected} & \textbf{Entropy} & \textbf{Immediate} & \textbf{Next GRPO} & \textbf{Final} \\
\midrule
0.65 & 100 & 0.69 & 0.438 & 0.432 & 0.434 \\
0.75 & 80 & 0.78 & 0.436 & 0.435 & 0.437 \\
0.85 & 60 & 0.86 & 0.434 & 0.437 & \textbf{0.440} \\
0.90 & 40 & 0.91 & 0.433 & 0.436 & 0.438 \\
0.95 & 20 & 0.95 & 0.431 & 0.434 & 0.435 \\
\bottomrule
\end{tabular}
\end{table}

The final column is single peaked.
A threshold of $0.65$ admits every checkpoint, so the rule reduces to selecting the best immediate validation accuracy and recovers the variant reported as w/o entropy aware handoff in Table~\ref{tab:ablation}.
A threshold of $0.95$ is the opposite failure and stops consolidation after twenty steps, discarding most of the benefit that the stage would have provided.
Between these, the final result varies by $0.6$ points across the range from $0.75$ to $0.90$, so the rule does not require precise tuning, but the two extremes cost $0.5$ and $0.6$ points respectively.
A threshold of $0.80$ selects the same checkpoint as $0.85$ in this round and is omitted from the table.

The immediate and final columns are ordered inversely over the top three rows, which is the central observation of Section~\ref{sec:evolvability} restated as a sweep.
Selecting on immediate accuracy alone is not merely suboptimal, it selects in the wrong direction, because within one consolidation round accuracy and diversity move against each other and only the second is consumed by the round that follows.

\clearpage
\section{Sensitivity to Rollout Sizes}
\label{app:rollout_sizes}

Two rollout sizes govern the loop, namely the GRPO group size $G$ and the number of rollouts $K$ sampled per visited question during mining.
Table~\ref{tab:rollout_sizes} varies each while holding the other at its default, under the full three cycle schedule with Qwen2.5-3B-Instruct.
Cost is the total wall clock of the run relative to the default configuration.
Coverage is the fraction of visited questions that supply a privileged reference after the first round.

\begin{table}[!htbp]
\centering
\caption{
Sensitivity to the GRPO group size $G$ and the mining size $K$ on Qwen2.5-3B-Instruct.
Cost is total wall clock relative to the default configuration.
Coverage is the fraction of visited questions with at least one successful rollout after the first round.
}
\label{tab:rollout_sizes}
\setlength{\tabcolsep}{6pt}
\footnotesize
\begin{tabular}{lccccc}
\toprule
\textbf{Configuration} & \textbf{Single} & \textbf{Multi} & \textbf{Avg.} & \textbf{Coverage} & \textbf{Cost} \\
\midrule
$G=3$ & 0.515 & 0.368 & 0.431 & -- & 0.72 \\
$G=5$ (Search-E1) & 0.520 & 0.380 & \textbf{0.440} & -- & 1.00 \\
$G=8$ & 0.521 & 0.381 & 0.441 & -- & 1.46 \\
\midrule
$K=3$ & 0.513 & 0.375 & 0.434 & 0.55 & 0.83 \\
$K=5$ (Search-E1) & 0.520 & 0.380 & \textbf{0.440} & 0.62 & 1.00 \\
$K=10$ & 0.522 & 0.382 & 0.442 & 0.71 & 1.42 \\
\bottomrule
\end{tabular}
\end{table}

Both parameters show clear returns up to the default and very little beyond it.
Reducing the group size to three costs $0.9$ points, since a group of three frequently contains either no successful trajectory or no unsuccessful one, in which case the normalized advantage carries no information and no reference is mined.
Increasing it to eight adds $0.1$ points for forty six percent more compute.
Reducing the mining size to three costs $0.6$ points and lowers reference coverage from $0.62$ to $0.55$, so a larger share of questions receives no consolidation signal in a given round.
Increasing it to ten raises coverage to $0.71$ but adds only $0.2$ points, because the additional references are drawn from questions the policy solves rarely, whose successful rollouts tend to be long and inefficient and are therefore poor candidates under the selection rule of Appendix~\ref{app:reference_selection}.
We keep $G=K=5$ throughout, which also matches the group size used by the outcome reward baselines and keeps the exploration stage identical to theirs.

\clearpage
\section{Search Behavior Across Rounds}
\label{app:behavior}

A consolidation stage that improves accuracy could in principle do so by changing how much the policy retrieves rather than how well it retrieves.
Two degenerate outcomes are possible, namely learning to issue more queries in the hope of covering the answer, and learning to skip retrieval and answer from parametric knowledge.
Table~\ref{tab:behavior} reports three behavioral quantities after every stage of the 3B Instruct run, measured on the evaluation benchmarks.

\begin{table}[!htbp]
\centering
\caption{
Search behavior after every stage of Search-E1 training with Qwen2.5-3B-Instruct, measured on the seven evaluation benchmarks.
Calls is the average number of search calls per rollout, Quality is the fraction of retrieval calls whose passages contain the gold answer, and Length is the average number of generated tokens per rollout.
}
\label{tab:behavior}
\setlength{\tabcolsep}{6pt}
\footnotesize
\begin{tabular}{lcccc}
\toprule
\textbf{Stage} & \textbf{Calls} & \textbf{Quality (\%)} & \textbf{Length} & \textbf{Avg. EM} \\
\midrule
GRPO$_1$ & 2.03 & 49.1 & 412 & 0.415 \\
OPSD$_1$ & 2.06 & 50.4 & 398 & 0.421 \\
GRPO$_2$ & 2.09 & 51.9 & 405 & 0.429 \\
OPSD$_2$ & 2.11 & 52.8 & 392 & 0.434 \\
GRPO$_3$ & 2.11 & 53.6 & 397 & 0.437 \\
OPSD$_3$ & 2.12 & 54.3 & 389 & 0.440 \\
\bottomrule
\end{tabular}
\end{table}

\begin{figure}[!htbp]
\centering
\includegraphics[width=0.55\linewidth]{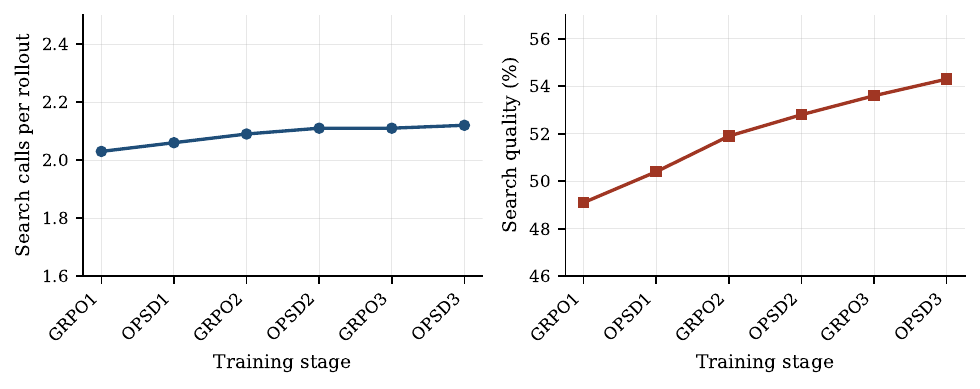}
\caption{
Search behavior across the stages of Search-E1 training with Qwen2.5-3B-Instruct.
The number of search calls per rollout stays nearly constant while the fraction of retrievals returning the gold answer rises monotonically, so the improvement comes from better queries rather than from more of them.
}
\label{fig:behavior}
\end{figure}

Neither degenerate outcome occurs.
The number of search calls per rollout moves from $2.03$ to $2.12$ over the entire schedule, a change of less than five percent, while search quality rises from $49.1$ to $54.3$ percent and the average Exact Match rises by $2.5$ points.
The improvement therefore comes from the content of the queries rather than from their number.
Generated length falls slightly, from $412$ to $389$ tokens, which is consistent with the selection rule favoring efficient references, but the change is small enough that it cannot account for the accuracy gain.
The consolidation stages contribute most of the movement in search quality, with an average rise of $1.0$ points per OPSD stage against $1.2$ points per GRPO stage, so both stages improve retrieval and neither is merely stabilizing the other.

\clearpage
\section{Robustness Across Random Seeds}
\label{app:seeds}

We repeat the complete three cycle schedule with three random seeds on Qwen2.5-3B-Instruct, together with the GRPO only control under identical settings.
Table~\ref{tab:seeds} reports the mean and standard deviation across seeds.

\begin{table}[!htbp]
\centering
\caption{
Exact Match mean and standard deviation across three random seeds on Qwen2.5-3B-Instruct.
The GRPO only control uses the same schedule, question recycling and evaluation protocol.
}
\label{tab:seeds}
\footnotesize
\setlength{\tabcolsep}{3pt}
\renewcommand{\arraystretch}{1.0}
\newcommand{\seedstat}[2]{\begin{tabular}[c]{@{}c@{}}$#1$\\[-2pt]{\scriptsize$\pm #2$}\end{tabular}}
\begin{tabular*}{\linewidth}{@{\extracolsep{\fill}}l ccc cccc c@{}}
\toprule
& \multicolumn{3}{c}{\textit{Single-Hop}} & \multicolumn{4}{c}{\textit{Multi-Hop}} & \\
\cmidrule(lr){2-4} \cmidrule(lr){5-8}
\textbf{Method} & NQ & TriviaQA & PopQA & HotpotQA & 2Wiki & MuSiQue & Bamboogle & \textbf{Avg.} \\
\midrule
Search-E1 & \seedstat{0.474}{0.007} & \seedstat{0.626}{0.005} & \seedstat{0.461}{0.008} & \seedstat{0.427}{0.010} & \seedstat{0.436}{0.012} & \seedstat{0.193}{0.009} & \seedstat{0.464}{0.024} & \seedstat{0.440}{0.006} \\
\addlinespace[3pt]
GRPO only & \seedstat{0.462}{0.008} & \seedstat{0.612}{0.006} & \seedstat{0.447}{0.009} & \seedstat{0.410}{0.011} & \seedstat{0.414}{0.013} & \seedstat{0.180}{0.010} & \seedstat{0.456}{0.026} & \seedstat{0.426}{0.007} \\
\bottomrule
\end{tabular*}
\end{table}

The per benchmark variability is uneven and follows the size of the evaluation split.
Standard deviations are smallest on the large single hop test sets and largest on Bamboogle, whose $125$ examples make a single question worth $0.8$ points.
The gap between the two methods exceeds one standard deviation on every benchmark except Bamboogle, and the averages are separated by roughly two standard deviations.
We describe the difference as consistent rather than as the outcome of a formal test, since three seeds is a small sample under reinforcement learning variance.
Single benchmark comparisons on Bamboogle should be read with particular caution throughout this paper, including the comparison against GiGPO in the main results tables.

\clearpage
\section{Case Study}
\label{app:case_study}

Figure~\ref{fig:case_study} follows one multi hop question through the schedule.
It shows the rollout produced after the first exploration round, the privileged skeleton mined for this question after the second exploration round, and the rollout produced by the final policy.
The example illustrates the behavior that the aggregate statistics of Appendix~\ref{app:behavior} describe, namely a change in the content of the queries at an unchanged number of retrievals.

\begin{figure}[!htbp]
\centering
\footnotesize
\fbox{\parbox{0.94\linewidth}{
\textbf{Question.} What is the date of death of Alexander of Masovia's father?\\
\textbf{Gold answer.} 21 January 1426\\[6pt]
\textit{After GRPO$_1$, incorrect}\\[2pt]
\texttt{<think>} I need the date of death of Alexander of Masovia's father. \texttt{</think>}\\
\texttt{<search>} Alexander of Masovia father date of death \texttt{</search>}\\
\texttt{<information>} [Alexander of Masovia] Alexander of Masovia (1400 to 2 June 1444) was the second son of Siemowit IV, Duke of Masovia. \texttt{</information>}\\
\texttt{<think>} The passage gives the dates 1400 and 1444. \texttt{</think>}\\
\texttt{<search>} Alexander of Masovia death \texttt{</search>}\\
\texttt{<information>} [Alexander of Masovia] ... died on 2 June 1444 in Trento ... \texttt{</information>}\\
\texttt{<answer>} 2 June 1444 \texttt{</answer>}\\[6pt]
\colorbox{yellow!25}{\parbox{0.97\linewidth}{
\textit{Privileged skeleton mined after GRPO$_2$, teacher only}\\[2pt]
\texttt{<search>} Alexander of Masovia father \texttt{</search>} $\rightarrow$ \texttt{<search>} Siemowit IV Duke of Masovia date of death \texttt{</search>}
}}\\[6pt]
\textit{After OPSD$_3$, correct}\\[2pt]
\texttt{<think>} First identify the father, then look up his death date. \texttt{</think>}\\
\texttt{<search>} Alexander of Masovia father \texttt{</search>}\\
\texttt{<information>} [Alexander of Masovia] ... was the second son of Siemowit IV, Duke of Masovia. \texttt{</information>}\\
\texttt{<think>} The father is Siemowit IV. Now I need his date of death. \texttt{</think>}\\
\texttt{<search>} Siemowit IV Duke of Masovia date of death \texttt{</search>}\\
\texttt{<information>} [Siemowit IV] Siemowit IV (c. 1352 to 21 January 1426) was Duke of Masovia. \texttt{</information>}\\
\texttt{<answer>} 21 January 1426 \texttt{</answer>}
}}
\caption{
One 2Wiki question at three points of the schedule.
The early policy receives the father's name in the first retrieval but does not use it in the second query, and reports Alexander's own date of death.
The privileged skeleton mined from a successful sibling rollout separates the two hops.
The final policy issues the same number of retrievals as the early one but resolves the bridge entity first, which is the behavior the skeleton encodes.
}
\label{fig:case_study}
\end{figure}

Both rollouts receive the father's name in the first retrieval, but only the successful rollout uses that bridge entity in its next query. The failure is therefore in using the returned information to guide the next search. The privileged skeleton supplies an ordered query sequence without retrieved passages or the final answer. This case illustrates the intended behavior of consolidation; it does not by itself establish that distillation caused the change.
Note that the final rollout issues two retrievals, exactly as the failed one did, which is the individual instance of the aggregate finding that consolidation changes query content at an unchanged retrieval budget.

\clearpage
\section{Prompts and Algorithm Details}
\label{app:prompts}

\paragraph{Inference prompt.}
The student and our internal GRPO controls use the same prompt in the original four-call setting. External baselines retain their reported prompts.

\begin{figure}[h]
\centering
\footnotesize
\fbox{\parbox{0.94\linewidth}{
\texttt{You are a helpful assistant. Answer the given question. You must}\\
\texttt{reason inside <think> and </think> first. If you lack knowledge,}\\
\texttt{you may call a search engine by writing a query inside <search>}\\
\texttt{and </search>, and the top passages will be returned inside}\\
\texttt{<information> and </information>. You may search as many times}\\
\texttt{as you need, up to four. When you know the answer, give it}\\
\texttt{inside <answer> and </answer> without any explanation.}\\[4pt]
\texttt{Question: \{question\}}
}}
\caption{Standard inference prompt $P_S(q)$, used by the student and at evaluation time.}
\label{fig:prompt_student}
\end{figure}

\paragraph{Privileged block.}
During consolidation the teacher receives one additional block, placed after the question and before the student rollout.
It contains the search spans of the reference in order, joined by an arrow, and is truncated to $512$ tokens.
No outcome label is attached, since only successful references are used.

\begin{figure}[h]
\centering
\footnotesize
\fbox{\parbox{0.94\linewidth}{
\texttt{[Reference search plan] A previous successful attempt at this}\\
\texttt{question issued the following queries in order.}\\
\texttt{<search> \{query 1\} </search> -> <search> \{query 2\} </search> -> ...}
}}
\caption{Privileged block $\mathcal{M}(\tau_q^{\mathrm{ref}})$, visible to the teacher only.}
\label{fig:prompt_teacher}
\end{figure}

\paragraph{Procedure.}
Figure~\ref{fig:algorithm} summarizes one complete cycle.

\begin{figure}[h]
\centering
\footnotesize
\fbox{\parbox{0.94\linewidth}{
\textbf{One Search-E1 cycle, starting from policy $\pi_r$ and pool $\mathcal{D}_r$}\\[4pt]
1. Run GRPO on $\pi_r$ for the scheduled number of steps using the outcome reward only, visiting a subset $\mathcal{V}_r \subseteq \mathcal{D}_r$, and obtain $\pi_r^{\mathrm{RL}}$.\\
2. For every $q \in \mathcal{V}_r$, sample $K$ rollouts from $\pi_r^{\mathrm{RL}}$ and record $c_r(q)$, the number of correct rollouts.\\
3. For every $q$ with $c_r(q) > 0$, select the successful rollout with the fewest search calls, break ties by skeleton length, and store $\mathcal{M}(\tau_q^{\mathrm{ref}})$.\\
4. Set $\mathcal{D}_{r+1} = \mathcal{D}_r \setminus \{q \in \mathcal{V}_r : c_r(q) = K\}$.\\
5. Initialize the teacher and the student from $\pi_r^{\mathrm{RL}}$ and freeze the teacher.\\
6. For each consolidation step, sample $\tau^{\mathrm{stu}}$ from the student under $P_S(q)$, evaluate the frozen teacher on $\tau^{\mathrm{stu}}$ under $P_T(q, \tau_q^{\mathrm{ref}})$, and update the student on the clipped forward KL of Equation~\ref{eq:opsd}. Store a checkpoint every twenty steps.\\
7. Among stored checkpoints whose entropy retention on the calibration set is at least $0.85$, select the one with the highest validation Exact Match and call it $\pi_{r+1}$.
}}
\caption{One cycle of exploration and consolidation. Reference answers determine correctness; no step introduces an external model, a reward model, or new annotations.}
\label{fig:algorithm}
\end{figure}

\clearpage
\begingroup
\section{Baseline Sources and Comparison Scope}
\label{app:baseline_sources}

Tables~\ref{tab:main_3b} and~\ref{tab:main_7b} compare reported performance for matched Qwen2.5 model sizes and Base/Instruct initializations in the QA retrieval setting.
They combine original reports, results reported by other papers, and our reproduction of AutoRefine-Instruct (7B).
Table~\ref{tab:baseline_sources} identifies these sources.
Our GRPO controls and Search-E1 ablations are trained in our own implementation.
The shared-configuration reruns in Appendix~\ref{app:controlled} are separate experiments.

\begin{table}[!htbp]
\centering
\caption{Sources of the baseline results in the main tables. Version numbers identify the reports from which the rows were taken.}
\label{tab:baseline_sources}
\footnotesize
\setlength{\tabcolsep}{4pt}
\renewcommand{\arraystretch}{1.15}
\begin{tabular}{@{}p{0.055\linewidth}p{0.30\linewidth}p{0.59\linewidth}@{}}
\toprule
Code & Rows & Source and result type \\
\midrule
A & Search-R1-Base / Instruct (3B) & AutoRefine (NeurIPS 2025; arXiv v5), Table 1; published results matching Search-R1's GRPO rows. \\
A & Search-R1-Base (7B) & AutoRefine, Table 3; the complete reported row, including its $0.408$ average. \\
R & Search-R1-Instruct (7B) & Search-R1 (COLM 2025; arXiv v5), Table 3; the GRPO row, replacing the PPO row in the original submission. \\
A & ReSearch-Base / Instruct (3B) & AutoRefine, Table 1; third-party reproductions using ReSearch's public code. \\
A & AutoRefine-Base (3B / 7B), Instruct (3B) & AutoRefine, Tables 1 and 3; original authors' results. \\
O & AutoRefine-Instruct (7B) & Our reproduction using Qwen2.5-7B-Instruct and AutoRefine's public code and supplied configuration. \\
S & MR-Search, StepSearch, Thinker & SD-Search (NeurIPS 2026; arXiv v1), Tables 1 and 2; published comparison results. Thinker rows are SFT results. StepSearch reports only the 4 multi-hop benchmarks. \\
S & SD-Search-Base / Instruct & SD-Search, Tables 1 and 2; original authors' results. \\
G & GiGPO-Instruct & GiGPO (NeurIPS 2025; arXiv v3), Table 2; original authors' results. \\
I & IG-Search-Base / Instruct & IG-Search (arXiv v1), Tables 1 and 3; original authors' results. \\
\bottomrule
\end{tabular}
\end{table}

We retain each complete reported row, keeping its benchmark scores and aggregate together.
The single-hop and multi-hop averages discussed here are arithmetic means of the 3 and 4 benchmark scores, respectively, rounded to 3 decimal places; the main tables retain each source's reported overall average.
For StepSearch, the single-hop and 7-benchmark averages remain unreported.
The highest-average claim refers to methods reporting all 7 benchmarks, rather than dominance on every dataset or task group.
In particular, among the Instruct variants, Search-E1 has the highest single-hop average at both scales and the highest multi-hop average at 3B; GiGPO has the higher multi-hop average at 7B.

\paragraph{Search-R1 variants.}
The original 7B Instruct row used the PPO result ($0.385$) rather than the stronger GRPO result ($0.396$).
Table~\ref{tab:main_7b} now uses the full GRPO row: $0.429$, $0.623$, $0.427$, $0.386$, $0.346$, $0.162$, and $0.400$ on NQ, TriviaQA, PopQA, HotpotQA, 2Wiki, MuSiQue, and Bamboogle, respectively.
Search-E1's $0.482$ average remains $8.6$ pp higher.
Table~\ref{tab:searchr1_variants} separately reports both Search-R1 variants from its original Table 3 for all 4 backbone configurations.
The $0.408$ Search-R1-Base (7B) row reported by AutoRefine remains identified separately in our main table.

\begin{table}[!htbp]
\centering
\caption{Average EM over 7 benchmarks for the original Search-R1 PPO and GRPO variants and Search-E1. Search-R1 values are from its COLM 2025 paper, arXiv v5, Table 3.}
\label{tab:searchr1_variants}
\footnotesize
\setlength{\tabcolsep}{10pt}
\begin{tabular}{lccc}
\toprule
Backbone & Search-R1 (PPO) & Search-R1 (GRPO) & Search-E1 \\
\midrule
Qwen2.5-3B-Base & 0.303 & 0.312 & 0.438 \\
Qwen2.5-3B-Instruct & 0.325 & 0.336 & 0.440 \\
Qwen2.5-7B-Base & 0.431 & 0.350 & 0.480 \\
Qwen2.5-7B-Instruct & 0.385 & 0.396 & 0.482 \\
\bottomrule
\end{tabular}
\end{table}

\paragraph{Retrieval and training settings.}
Our original setup follows AutoRefine in using NQ and HotpotQA for training, E5-base-v2 over December 2018 Wikipedia, 3 retrieved passages per query, and normalized EM on the corresponding benchmark splits.
This common QA retrieval setting does not imply identical training configurations.
For example, reported search limits include 3 calls for SD-Search and GiGPO and 5 for AutoRefine and IG-Search, compared with our original limit of 4.
The main-table results retain their source configurations, including the AutoRefine configuration used in our 7B Instruct reproduction.
Appendix~\ref{app:controlled} compares methods retrained with a shared 4-call limit, and Appendix~\ref{app:evaluation_budget} changes the evaluation limit for those fixed policies.

\paragraph{Interpreting task-group averages.}
Search-E1 scores higher than GiGPO on 3 of the 4 multi-hop datasets at both scales, whereas GiGPO has a large advantage on Bamboogle ($0.641$ at 3B and $0.689$ at 7B).
Bamboogle has 125 questions but receives the same weight as each other multi-hop benchmark.
Its contribution therefore helps explain why GiGPO has the higher 7B multi-hop average.
The controlled reruns below show the same task-group ranking under a shared setup.
The query-position ablation (Appendix~\ref{app:objective}) and search-quality analysis (Appendix~\ref{app:behavior}) support learning useful search behavior, while the case study (Appendix~\ref{app:case_study}) illustrates use of a returned bridge entity in a later query.
These observations do not imply that OPSD solves questions for which the current policy never produces a successful reference.

\clearpage
\section{Additional Experiments}
\label{app:additional_experiments}

This appendix reports additional controls and extensions of the QA evaluation.
Unless a table specifies a different scope, single-hop EM averages NQ, TriviaQA, and PopQA; multi-hop EM averages HotpotQA, 2WikiMultihopQA, MuSiQue, and Bamboogle; and overall EM averages all 7 benchmarks.
We first compute EM over individual questions within each benchmark, then average across benchmarks.
New training runs, evaluation-only changes to fixed policies, and reference results from the original experiments are identified separately.

\subsection{Controlled Comparisons at 7B}
\label{app:controlled}

We retrain GiGPO, IG-Search, and Search-E1 from the same Qwen2.5-7B-Instruct checkpoint, using the same NQ and HotpotQA training data, Wikipedia retrieval environment, hardware, software, and evaluation setup.
All 3 methods use a limit of 4 search calls during training and evaluation and 500 executed optimizer steps, including all GRPO and OPSD updates for Search-E1.
Evaluation uses a shared answer extractor and scorer.
Table~\ref{tab:controlled_7b} reports means over 3 training seeds and standard deviations for overall EM.
Matching optimizer steps does not equate the wall-clock cost of the methods; Appendix~\ref{app:budget} accounts separately for Search-E1's mining and distillation costs in the 3B setting.

\begin{table}[!htbp]
\centering
\caption{Controlled 7B comparison under a shared training and evaluation setup. Overall EM is the mean across all 7 benchmarks, not the unweighted mean of the 2 task-group columns.}
\label{tab:controlled_7b}
\footnotesize
\setlength{\tabcolsep}{12pt}
\begin{tabular}{lccc}
\toprule
Method & Single-hop EM & Multi-hop EM & Overall EM \\
\midrule
GiGPO-Instruct & 0.526 & 0.430 & $0.471 \pm 0.005$ \\
IG-Search-Instruct & 0.549 & 0.424 & $0.478 \pm 0.004$ \\
Search-E1-Instruct & 0.558 & 0.423 & $0.481 \pm 0.004$ \\
\bottomrule
\end{tabular}
\end{table}

Search-E1 has the highest mean overall EM in this comparison, while GiGPO retains the higher multi-hop average.
The overall differences are small relative to training variability; we report the observed ranking without claiming statistical significance.
Our internal 3B controls provide complementary evidence: Appendix~\ref{app:seeds} reports $0.440\pm0.006$ for Search-E1 versus $0.426\pm0.007$ for GRPO-only, and the 700-step GRPO control in Appendix~\ref{app:budget} peaks at $0.427$ and finishes at $0.422$.

\subsection{Evaluation Search Limits for Fixed Policies}
\label{app:evaluation_budget}

For each method, we keep the checkpoints from the 4-call training runs in Appendix~\ref{app:controlled} fixed and evaluate with limits of 3, 4, 5, and 8 search calls.
The retrieval index, decoding settings, evaluation seed set, and scoring code are unchanged.
The limit counts actual retrieval calls, and each policy can still generate a final answer after reaching it.
Table~\ref{tab:eval_budget} reports mean EM over the same 3 training seeds on the 4 multi-hop benchmarks.
The 4-call column repeats Table~\ref{tab:controlled_7b}.

\begin{table}[!htbp]
\centering
\caption{Multi-hop EM when changing only the evaluation search limit for policies trained with 4 calls.}
\label{tab:eval_budget}
\footnotesize
\setlength{\tabcolsep}{12pt}
\begin{tabular}{lcccc}
\toprule
Method & 3 calls & 4 calls & 5 calls & 8 calls \\
\midrule
GiGPO-Instruct & 0.426 & 0.430 & 0.431 & 0.431 \\
IG-Search-Instruct & 0.419 & 0.424 & 0.425 & 0.426 \\
Search-E1-Instruct & 0.419 & 0.423 & 0.424 & 0.425 \\
\bottomrule
\end{tabular}
\end{table}

\begin{table}[!htbp]
\centering
\caption{Search use under the 8-call evaluation limit. Percentages are computed per benchmark over all rollouts, including incorrect answers, and then averaged over the 4 multi-hop benchmarks.}
\label{tab:eval_budget_usage}
\footnotesize
\setlength{\tabcolsep}{10pt}
\begin{tabular}{lcc}
\toprule
Method & Answered within 3 calls (\%) & Used more than 4 calls (\%) \\
\midrule
GiGPO-Instruct & 82.4 & 4.3 \\
IG-Search-Instruct & 78.8 & 6.8 \\
Search-E1-Instruct & 80.6 & 5.9 \\
\bottomrule
\end{tabular}
\end{table}

``Answered within 3 calls'' means the policy produced its final answer after at most 3 retrieval calls.
Under the 8-call limit, $78.8$--$82.4\%$ of rollouts meet this condition.
Increasing the evaluation limit from 4 to 8 calls changes multi-hop EM by $+0.14$, $+0.18$, and $+0.16$ pp for GiGPO, IG-Search, and Search-E1, respectively, with differences calculated before rounding EM to 3 decimal places.
The multi-hop ranking is unchanged, including GiGPO's advantage.
This result concerns fixed policies trained with 4 calls; it does not measure learning under a larger training budget, which is tested in Appendix~\ref{app:larger_budget}.

\subsection{GRPO with the Same Trajectory Preference}
\label{app:ranking_reward}

This experiment tests whether rewarding the same preference used in reference selection can recover the benefit of OPSD without token-level distillation.
For each question, we use a group of 5 rollouts and rank the successful rollouts by search count, with fewer calls preferred, breaking ties by search-skeleton length.
Let $M_q$ denote the number of distinct (search count, skeleton length) pairs among successful rollouts, and let $k_i$ be rollout $i$'s rank starting from 0 for the best pair.
Rollouts tied on both criteria have the same rank.
With $c_i=\mathrm{EM}(\hat a_i,a)$ and $\lambda>0$, we use
\begin{equation}
r_i =
\begin{cases}
0, & c_i=0,\\
1+\lambda\left(1-\dfrac{k_i}{\max(M_q-1,1)}\right), & c_i=1.
\end{cases}
\label{eq:ranking_reward}
\end{equation}
Incorrect rollouts receive 0, and correct rollouts receive a reward between 1 and $1+\lambda$.
Search count takes priority over skeleton length.
For any group containing a correct answer, the highest-reward trajectories are exactly those eligible for selection as the Search-E1 reference.
If all rollouts are incorrect, all rewards are 0.
This preserves the correctness-first preference without changing the question-recycling criterion, which remains based on answer correctness.

We use Qwen2.5-3B-Instruct with Wikipedia retrieval and a limit of 4 search calls during training and evaluation.
Each reward-shaping run uses 700 optimizer steps, matching Search-E1's executed optimizer-step budget and the extended GRPO control in Appendix~\ref{app:budget}.
The GRPO controls share the initial checkpoint, training data, question-recycling rule, hardware, software, and decoding settings.
We sweep $\lambda\in\{0.05,0.10,0.20\}$ and select the coefficient and checkpoints using the held-out validation set.

\begin{table}[!htbp]
\centering
\caption{GRPO with a ranking reward aligned with reference selection. Overall EM and search calls cover all 7 benchmarks; multi-hop EM covers the 4 multi-hop benchmarks. Dashes mark values not reported for the extended GRPO control.}
\label{tab:ranking_reward}
\footnotesize
\setlength{\tabcolsep}{6pt}
\begin{tabular}{lcccc}
\toprule
Method & $\lambda$ & Overall EM & Multi-hop EM & Mean search calls \\
\midrule
GRPO-only (peak within 700 steps) & 0 & 0.427 & -- & -- \\
GRPO + ranking reward & 0.05 & 0.430 & 0.369 & 2.05 \\
GRPO + ranking reward & 0.10 & 0.433 & 0.372 & 1.98 \\
GRPO + ranking reward & 0.20 & 0.431 & 0.370 & 1.84 \\
Search-E1 & 0 & 0.440 & 0.380 & 2.12 \\
\bottomrule
\end{tabular}
\end{table}

The GRPO-only row is the $0.427$ peak in Appendix~\ref{app:budget}, rather than the $0.426$ obtained under the original GRPO schedule in Table~\ref{tab:ablation}.
With the validation-selected coefficient $\lambda=0.10$, the ranking reward reaches $0.433$ EM with $1.98$ searches per rollout.
Increasing $\lambda$ to $0.20$ reduces searches to $1.84$, while EM is $0.431$.
Search-E1 reaches $0.440$ EM with $2.12$ searches, a gain of $0.7$ pp overall and $0.8$ pp on multi-hop QA over the selected reward-shaping baseline.

The ranking reward improves over GRPO-only, while Search-E1 retains a gain under the same trajectory preference.
In GRPO, the preference becomes a group-normalized advantage shared across the policy-generated positions of the trajectory.
OPSD instead supplies a next-token distribution conditioned on the successful query sequence and the current student prefix.
This comparison supports the value of that token-level supervision; it does not isolate every difference between reinforcement learning and distillation.
The query-order ablation in Appendix~\ref{app:privileged_context} provides complementary evidence: shuffling the reference queries without changing their content or number reduces EM from $0.440$ to $0.431$.

\clearpage
\subsection{Larger Training and Evaluation Search Budgets}
\label{app:larger_budget}

We train both Search-E1 and GRPO-only with an 8-call limit using Qwen2.5-3B-Instruct and Wikipedia retrieval, then evaluate with limits of 8 and 12 calls.
Table~\ref{tab:larger_budget} includes the original 4-call results from Table~\ref{tab:ablation} for reference.
Unlike Appendix~\ref{app:evaluation_budget}, these runs change the training search limit as well as the evaluation limit.

\begin{table}[!htbp]
\centering
\caption{Larger search budgets on Qwen2.5-3B-Instruct. EM is averaged over the 4 multi-hop benchmarks.}
\label{tab:larger_budget}
\footnotesize
\setlength{\tabcolsep}{12pt}
\begin{tabular}{cccc}
\toprule
Training limit & Evaluation limit & GRPO-only EM & Search-E1 EM \\
\midrule
4 (original) & 4 & 0.365 & 0.380 \\
8 & 8 & 0.371 & 0.389 \\
8 & 12 & 0.373 & 0.390 \\
\bottomrule
\end{tabular}
\end{table}

With evaluation limits of 8 and 12, $20.8\%$ and $22.3\%$ of Search-E1 rollouts use more than 4 searches, respectively.
The evaluation therefore includes search sequences beyond the original limit.
Increasing both the training and evaluation limits from 4 to 8 raises Search-E1's multi-hop EM from $0.380$ to $0.389$.
Search-E1 retains a gain over GRPO-only at both larger evaluation limits.
Generalization to much longer sequences of dependent searches remains outside the evaluated scope.

\subsection{Training and Evaluation with Web Search}
\label{app:web_search}

We train and evaluate both methods using Google Search through the Serper API.
Each query returns webpage titles, URLs, and text snippets, which serve as the policy's retrieval observations.
For each method, the Wikipedia and Google Search runs start from the same Qwen2.5-3B-Instruct checkpoint and use the same training data, optimization schedule, search-call limits, and decoding settings.
Each run uses its corresponding retrieval backend throughout training and evaluation.
The backend comparison uses the 2 rows with an evaluation limit of 8 in Table~\ref{tab:web_search}; the original 4-call row is included for reference.

\begin{table}[!htbp]
\centering
\caption{Retrieval backend comparison on Qwen2.5-3B-Instruct. EM averages the 4 multi-hop benchmarks. The backend is used in both training and evaluation.}
\label{tab:web_search}
\footnotesize
\setlength{\tabcolsep}{6pt}
\begin{tabular}{lcccc}
\toprule
Retrieval backend & Training limit & Evaluation limit & GRPO-only EM & Search-E1 EM \\
\midrule
Wikipedia (original) & 4 & 4 & 0.365 & 0.380 \\
Wikipedia & 4 & 8 & 0.367 & 0.382 \\
Google Search (Serper) & 4 & 8 & 0.372 & 0.389 \\
\bottomrule
\end{tabular}
\end{table}

Search-E1 reaches $0.389$ multi-hop EM with Google Search, compared with $0.382$ using Wikipedia under the same call limits, an increase of $0.7$ pp.
It retains a $1.7$ pp gain over GRPO-only with Google Search.
This extends the evidence to a web retrieval backend used throughout training and evaluation, while retaining the QA task and answer-level supervision.

\clearpage
\subsection{Additional Model Backbones}
\label{app:backbones}

We train Search-E1 and GRPO-only on Qwen3-4B, Qwen3-8B, and Llama-3.1-8B-Instruct under the Wikipedia setup.
Within each pair, the methods share the initial checkpoint, training data, retrieval environment, search limit, and sampling settings.
Table~\ref{tab:additional_backbones} reports mean EM and standard deviation over 3 training seeds, averaged over all 7 benchmarks.

\begin{table}[!htbp]
\centering
\caption{Additional backbones under the Wikipedia retrieval setup. Values are 7-benchmark average EM, with standard deviations over 3 training seeds.}
\label{tab:additional_backbones}
\footnotesize
\setlength{\tabcolsep}{10pt}
\begin{tabular}{lccc}
\toprule
Backbone & GRPO-only EM & Search-E1 EM & Gain (pp) \\
\midrule
Qwen3-4B & $0.453\pm0.005$ & $0.469\pm0.004$ & $+1.6$ \\
Qwen3-8B & $0.491\pm0.004$ & $0.505\pm0.003$ & $+1.4$ \\
Llama-3.1-8B-Instruct & $0.421\pm0.006$ & $0.434\pm0.005$ & $+1.3$ \\
\bottomrule
\end{tabular}
\end{table}

Search-E1 improves mean EM on all 3 additional backbones, extending the evidence to a newer Qwen generation and another model family within QA.
The experiments in this appendix do not remove the need for outcome supervision: reference answers determine GRPO rewards and successful-reference selection, while the policy supplies the privileged search skeletons.
The teacher is a frozen copy of the policy at the beginning of each OPSD round, and its privileged context excludes the reference trajectory's retrieved passages and final answer.
Reference answers and privileged context are not needed at inference.
Tasks without reference answers would require another reliable outcome verifier; other task families would also require a suitable representation of successful action sequences.
\clearpage
\endgroup

\end{document}